\documentclass{article} 
\usepackage[preprint]{colm2025_conference}

\usepackage{microtype}
\usepackage{hyperref}
\usepackage{url}
\usepackage{booktabs}
\usepackage{censor}

\usepackage{lineno}
\usepackage{graphicx}
\usepackage{amssymb}
\usepackage{bm}

\usepackage{framed}

\definecolor{darkblue}{rgb}{0, 0, 0.5}
\hypersetup{colorlinks=true, citecolor=darkblue, linkcolor=darkblue, urlcolor=darkblue}

\title{Modifying Large Language Model Post-Training for Diverse Creative Writing}

\author{
\textbf{John Joon Young Chung$^1$\thanks{Correspondence: jchung@midjourney.com}, Vishakh Padmakumar$^2$,  Melissa Roemmele$^1$,} \\\textbf{Yuqian Sun$^1$ \& Max Kreminski$^1$} \\
$^1$Midjourney \\
$^2$New York University
}

\def\thickhline{\noalign{\hrule height1pt}}

\begin{document}

\ifcolmsubmission
\linenumbers
\fi

\maketitle

\begin{abstract}

As creative writing tasks do not have singular correct answers, large language models (LLMs) trained to perform these tasks should be able to generate diverse valid writings. However, LLM post-training often focuses on improving generation quality but neglects to facilitate output diversity. Hence, in creative writing generation, we investigate post-training approaches to promote both output diversity and quality. Our core idea is to include deviation---the degree of difference between a training sample and all other samples with the same prompt---in the training objective to facilitate learning from rare high-quality instances. By adopting our approach to direct preference optimization (DPO) and odds ratio preference optimization (ORPO), we demonstrate that we can promote the output diversity of trained models while minimally decreasing quality. Our best model with 8B parameters could achieve on-par diversity as a human-created dataset while having output quality similar to the best instruction-tuned models we examined, GPT-4o and DeepSeek-R1. We further validate our approaches with a human evaluation, an ablation, and a comparison to an existing diversification approach, DivPO.\makeatletter
\footnote{
\@ifpackagewith{colm2025_conference}{submission}{We provide our code in [redacted for a review---refer to supplementary materials].}{We provide our code in \url{https://github.com/mj-storytelling/DiversityTuning}.}
}
\makeatother
\end{abstract}

\section{Introduction}

In creative writing, there is no single ``gold'' answer but multiple valid ways that the writing can unfold~\citep{flower1981cognitive}. 
For example, given a writing prompt "write a story about a dog on the moon," many different stories might be written in response, with the focus ranging from the dog's adventure to the dog's lonely moon life.
This kind of divergent thinking ability has also been stressed as one aspect of creative intelligence~\citep{runco2012divergent, guilford1957creative}. Hence, when training large language models (LLMs) to generate creative writing, these models should learn to find and consider diverse paths and endings.
While various post-training approaches---such as proximal policy optimization (PPO)~\citep{ouyang2022training} or direct preference optimization (DPO)~\citep{rafailov2023direct}---can strongly increase the quality of LLM output, this tuning also seems to decrease output diversity~\citep{padmakumar2024does, anderson2024homogenization, kirk2024understanding, xu2024echoesaiquantifyinglack}. Low output diversity can cause issues in creative task contexts—it can lead users of LLM assistants to produce homogenous content~\citep{anderson2024homogenization, padmakumar2024does} or show a limited set of biased outputs to the user despite the existence of various valid responses~\citep{venkit2024searchenginesaiera, venkit2023unmasking, jakesch2023cowriting}. 

Previous research investigated ways to increase LLM output diversity, but has largely focused on how to make the best use of already post-trained models, e.g, by prompting~\citep{wong2024simplestratdiversifyinglanguagemodel, hayati2024far} or adjusting sampling temperature~\citep{chung2023increasing}. 
While a few works studied tuning the model to facilitate generation diversity, they targeted a narrow set of simple tasks (e.g., baby name generation)~\citep{zhang2024forcing}.

\begin{figure}[t]
\begin{center}
    \includegraphics[width=\textwidth]{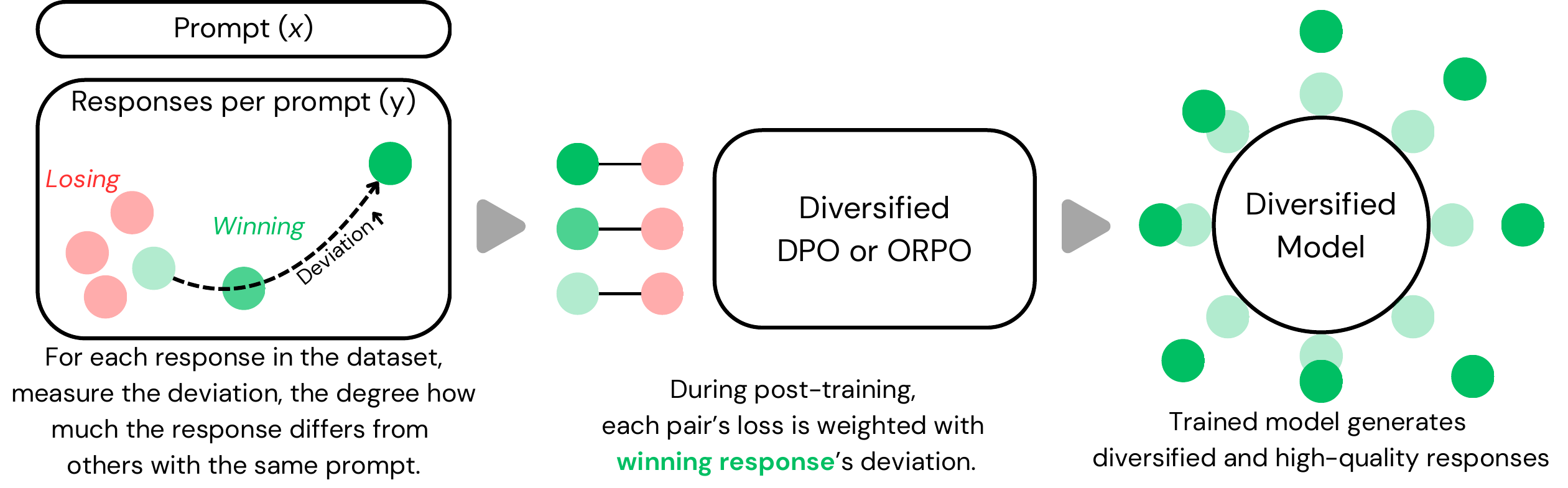}
\end{center}
\caption{Our post-training approach to diversify creative writing generation while maintaining quality.}
\label{fig:teaser}
\end{figure}

For creative writing generation, we explore post-training approaches to facilitate output diversity while maintaining quality. We propose to consider diversity as a part of training objectives. Specifically, we factor in \textit{deviation}, a measure of how much a training instance differs from other candidate samples for the same prompt. 
We incorporate deviation into different LLM post-training methods, introducing diversified versions of DPO and odds ratio preference optimization (ORPO)~\citep{hong2024orpomonolithicpreferenceoptimization}.

Our results demonstrate that diversified DPO and ORPO could facilitate semantic and style diversity (defined in Section~\ref{sec:diversity_metric}) while minimally decreasing the quality of writing. Our trained models had higher diversity than existing instruction-tuned models, such as GPT-4o~\citep{gpt4o}, Claude-3.5-Sonnet~\citep{claude3.5sonnet}, or DeepSeek-R1~\citep{deepseekai2025deepseekr1incentivizingreasoningcapability}. The writing quality of our best model (a \texttt{Llama-3.1-8B}-based diversified DPO model) was on par with that of existing best-quality models while having a similar level of diversity to a dataset created by humans. 
Human evaluation that compares diversified DPO, original DPO, and GPT-4o further validates that our approach could generate diverse output while maintaining output quality.
With an additional experiment that varies the maximum number of training instances per prompt, we show that, except when there are too few instances per prompt, our approach robustly increases output diversity while having on-par quality with the non-diversified approach and a contemporary work, DivPO~\citep{lanchantin2025diversepreferenceoptimization}. 
With too few instances per prompt, the quality drops with our approach. However, we demonstrate that this quality issue can be fixed by slightly tweaking the objective function or training with high-quality instances.
\section{Related Work}

While LLM post-training approaches, including PPO~\citep{ouyang2022training}, DPO~\citep{rafailov2023direct}, ORPO~\citep{hong2024orpomonolithicpreferenceoptimization}, have improved the model's instruction following capability and output quality, these often resulted in low output diversity~\citep{kirk2024understanding, go2023aligning, casper2023open}. 
Researchers observed such decreased diversity in creative writing generation, with similar narrative elements echoed repetitively in multiple generations~\citep{xu2024echoesaiquantifyinglack}.
Low diversity could be detrimental in creative task settings. For example, researchers found that people's creative products can homogenize when they use post-trained LLMs~\citep{anderson2024homogenization, padmakumar2024does, chen2025understandingdesignfixationgenerative}. 

A way to facilitate output diversity is adjusting decoding approaches~\citep{ippolito2019comparison}. For beam search, researchers proposed penalizing siblings~\citep{li2016simplefastdiversedecoding} or counting in diversity during search~\citep{vijayakumar2017diverse}.
For sampling, researchers explored adjusting temperatures~\citep{tevet2021evaluating, chung2023increasing}.
However, these often have quality-diversity trade-offs~\citep{chung2023increasing, tevet2021evaluating, zhang2021trading}, limiting the extent of usage. To alleviate it, researchers introduced top-k~\citep{fan2018hierarchical}, top-p~\citep{Holtzman2020The}, and min-p~\citep{minh2025turning} sampling, cutting out the least probable tokens during decoding. Researchers also introduced adaptive temperature to balance quality and diversity~\citep{zhang2024edtimprovinglargelanguage}.
With improving instruction-following capability, researchers explored prompting-based diversification, such as self-critiquing for diversity~\citep{lahoti2023improving}, 
iteratively prompting to avoid previous generations~\citep{hayati2024far}, and 
guiding LLM generation with answer space dimensions~\citep{wong2024simplestratdiversifyinglanguagemodel, suh2024luminate}, answer set programming~\citep{wang2024guidingdiversifyingllmbasedstory}, or evolutionary algorithms~\citep{bradley2024qualitydiversity}. 
Only few, however, investigated tuning LLMs for diversity. \cite{zhang2024forcing} is one, but it deals with simple tasks of baby name or random number generation. 
DivPo~\citep{lanchantin2025diversepreferenceoptimization} is another, which filters the preference dataset to have highly diverse winning instances and lowly diverse losing instances. 
While it facilitated diversity, it still showed trade-offs between the quality and diversity of outputs. 
To diversify creative writing generation while preserving quality, we leverage deviation---how a training instance differs from other instances---as a part of the learning objective. 
\section{Preliminaries}

We start with the problem setting of creative writing generation and the metrics. Then, we explain two post-training approaches we extended: DPO and ORPO.\footnote{We did not experiment with a seminal post-training approach, PPO. PPO is known to be technically difficult and complex to train~\citep{casper2023open}. 
As having a robust reward model for creative writing is difficult due to subjectivity in evaluation, it was more of the case in our context.
Unlike other approaches, our pilots on PPO did not pan out in the examined creative writing dataset. }

\subsection{Problem Setting and Metrics}

For creative writing tasks, given a prompt ($x$), LLMs ($\bm{\theta}$) should generate diverse, high-quality outputs ($y_i)$ so that end-users can get various valid options. Here, we define ``valid'' as whether the generated output satisfies the prompt. For instance, if a user inputs a prompt to discover possible next events in their story, showing diverse branches would be desirable. In this setting, we formulate quality and diversity as follows:

\begin{equation}
    \bm{\mathrm{Quality}}(x, \bm{\theta})=\mathbb{E}^N_{i=1} \left[ \bm{\mathrm{r}}(x, y_i) \right] \quad \mathrm{where} \quad y_i \sim \bm{p}_{\bm{\theta}}(y|x)
\end{equation}
\begin{equation}
    \bm{\mathrm{Diversity}}(x, \bm{\theta})=\bm{\mathrm{Diversity}_y}(Y) \quad \mathrm{where} \quad Y=\{y_i\}_{i=1}^N
\end{equation}

The above assumes sampling $N$ responses from $x$.
$\bm{\mathrm{r}}$ calculates a reward (i.e., validity and quality of a response) over a prompt-response pair, and $\bm{\mathrm{Diversity}_y}$ evaluates diversity of all sampled responses.
While many options exist for $\bm{\mathrm{Diversity}_y}$~\citep{cox2021directeddiversity}, we use mean pairwise distance between a set of outputs:

\begin{equation}
    \bm{\mathrm{Diversity}_y}(Y) = \mathbb{E}^N_{i=1} \left[ \mathbb{E}^{N-1}_{j=1, j\neq i} \left[ \bm{d}(y_i, y_j) \right] \right] \quad \mathrm{where} \quad Y=\{y_i\}_{i=1}^N
\end{equation}

Note that $\bm{d}$ is a distance function and $\mathbb{E}^{N-1}_{j=1, j\neq i} \left[ \bm{d}(y_i, y_j) \right]$ indicates deviation, how $y_i$ is different from all other samples.

\subsection{Post-training Approaches}

As post-training could be noisy with limited samples, some post training methods like DPO require supervised fine-tuning (SFT)  
to guide the trained models to have desired generation behaviors.
Starting from pre-trained models, the SFT model can be trained by maximizing the likelihood of response $y$ given prompt $x$ in dataset $\mathcal{D}$ with the below loss:

\begin{equation}
    \mathcal{L}_{\mathrm{SFT}} = - \mathbb{E}_{(x, y) \in \mathcal{D}} \left[ \mathrm{log} \ \bm{p_{\theta}}(y|x) \right]
\end{equation}

\subsubsection{Direct Preference Optimization}

DPO is an approach that optimizes the policy model directly on the dataset of $(x, y^w, y^l)$, increasing the likelihood of $y^w$ over $y^l$. With the ratio between the likelihood of the policy model and the reference SFT model as an implicit reward, the training objective is as follows:

\begin{equation}
    \mathcal{L}_{\mathrm{DPO}} = -\mathbb{E}_{(x, y^w, y^l) \in \mathcal{D}} \left[ 
        \mathrm{log} \: \sigma 
            (
                \beta \: \mathrm{log} \frac{\bm{p_{\theta}}(y^w|x)}{\bm{p_{\mathrm{SFT}}}(y^w|x)}
                \: - \:
                \beta \: \mathrm{log} \frac{\bm{p_{\theta}}(y^l|x)}{\bm{p_{\mathrm{SFT}}}(y^l|x)}
            )
    \right]
    \label{eq:dpo}
\end{equation}

\subsubsection{Odds Ratio Preference Optimization}

ORPO is another approach that directly optimizes over $(x, y^w, y^l)$. However, for signaling preferences, it does not use an SFT model as a reference. Instead, it uses the odds ratio as a preference signal to model training:

\begin{equation}
    \bm{\mathrm{OR}_\theta} (y^w, y^l) = \frac{\bm{\mathrm{odds}_\theta}(y^w|x)}{\bm{\mathrm{odds}_\theta}(y^l|x)}
    \quad
    \mathrm{where}
    \quad
    \bm{\mathrm{odds}_\theta}(y|x) = \frac{\bm{p_\theta}(y|x)}{1 - \bm{p_\theta}(y|x)}
\end{equation}

Specifically, the training objective combines log-likelihood (over the winning responses) and log odds ratio as below. Note that ORPO starts training from a base model, not an SFT model.

\begin{equation}
    \mathcal{L}_\mathrm{ORPO} = -\mathbb{E}_{(x, y^w, y^l) \in \mathcal{D}} \left[ 
        \mathrm{log} \: \bm{p_\theta}(y^w|x) 
        +
        \lambda \: \mathrm{log} \: \sigma (\mathrm{log} \: \frac{\bm{\mathrm{odds}_\theta}(y^w|x)}{\bm{\mathrm{odds}_\theta}(y^l|x)})
    \right]
\end{equation}
\section{Diversified DPO and ORPO}

We introduce an approach to promote both quality and diversity in LLM post-training. The main idea is to factor in \textit{deviation} ($\delta$) into the objective function. We define deviation as ``the degree of how much a training instance differs from all other instances with the same prompt.`` By factoring this in, we aim to increase the likelihood of models generating output of high quality that deviates from "typical" outputs. 
Note that we assume there are enough prompts with more than three responses, as the concept of ``deviation'' holds only for this case (i.e., with two responses, deviations would be the same for both of them). 

\paragraph{Diversified DPO (DDPO)}
To count deviation into DPO, for each pair of winning-losing responses, we weighted the pair's training objective with the deviation of the winning instance. As the pair's training objective is to learn the winning response's behavior, this weighting would emphasize rare winning instances more than common winning ones. 

\begin{equation}
    \mathcal{L}_{\mathrm{DDPO}} = -\mathbb{E}_{(x, y^w, y^l) \in \mathcal{D}} \left[ 
        \delta^w \: \mathrm{log} \: \sigma 
            (
                \beta \: \mathrm{log} \frac{\bm{p_{\theta}}(y^w|x)}{\bm{p_{\mathrm{SFT}}}(y^w|x)}
                \: - \:
                \beta \: \mathrm{log} \frac{\bm{p_{\theta}}(y^l|x)}{\bm{p_{\mathrm{SFT}}}(y^l|x)}
            )
    \right]\footnote{\label{ft:deviation}We explain how we quantify $\delta$ in Section~\ref{sec:deviation_measure} and~\ref{app:deviation_transform}.}
    \label{eq:ddpo}
\end{equation}

\paragraph{Diversified ORPO (DORPO)}
We extended ORPO similarly, by weighting the pair loss with deviation of the winning instances. Here, note that we weighted both log-likelihood and log odds ratio terms, as both contribute to learning the behavior of winning response. 

\begin{equation}
    \mathcal{L}_\mathrm{DORPO} = -\mathbb{E}_{(x, y^w, y^l) \in \mathcal{D}} \left[ 
    \delta^w \: 
        \mathrm{log} \:  \bm{p_\theta}(y^w|x) 
        +
        \lambda \: {\delta^w} \: \mathrm{log} \: \sigma (\mathrm{log} \: \frac{\bm{\mathrm{odds}_\theta}(y^w|x)}{\bm{\mathrm{odds}_\theta}(y^l|x)})
    \right]\textsuperscript{\ref{ft:deviation}}
\end{equation}

\section{Experiments}
\label{sec:experiments}

\subsection{Experiment Settings and Metrics}

\paragraph{Dataset} We focus on creative writing by using r/writingPrompts dataset~\citep{fan2018hierarchical}\footnote{\label{ft:hgdataset}\url{https://huggingface.co/datasets/euclaise/WritingPrompts_preferences}}. The data originates from r/writingPrompts subreddit, where users post writing prompts ($x$) and other users share their creative writing to the prompts as comments ($y$). Hence, there can be multiple writings for a single prompt, which we could leverage to train a model that generates diverse outputs to the same prompt. Moreover, each writing has a user upvote score, $s$, which we used as a signal for the writing quality. We use these scores to 1) train a reward model for evaluation and 2) craft a binary preference dataset with pairs of winning and losing responses to train generation models.\footnote{Upvote scores are somewhat noisy reward signals. For example, some decent writings might have gotten low attention and received similar upvotes compared to lower-quality creative writings.} While we split the data into train and test sets (421330 and 45868 prompt-response pairs, respectively), we describe data processing details in Appendix~\ref{app:dataprocessing}.

\paragraph{Evaluation Task and Metrics} 
\label{sec:diversity_metric}
To evaluate the quality and diversity, we sampled four instances per each evaluation prompt. Then, we measured the quality of each sample and the diversity between those four instances. We used 1000 evaluation prompts from the test dataset, resulting in 4000 samples. We detail our sampling approach in Appendix~\ref{app:eval_details}.

We automatically evaluated the model output's quality and diversity. For the output quality  (\texttt{reddit-reward}), we trained a reward model out of $(x, y, s)$ triplets in the dataset as a regression model that predicts $s$ from $(x, y)$. We describe the training details and performances of the reward model in Appendix~\ref{app:rewardmodel}. 

For diversity evaluation, we embedded sampled $y$ with embedding models and measured mean pairwise cosine distances between all samples from the same prompt. We focused on 1) \texttt{semantic diversity} and 2) \texttt{style diversity}. For embedding models, we used \texttt{jinaai/jina-embeddings-v3}~\citep{sturua2024jinaembeddingsv3multilingualembeddingstask} and \texttt{AnnaWegmann/Style-Embedding}~\citep{wegmann-etal-2022-author}, respectively for semantic and style diversity. Note that style embeddings capture whether the same or different people would have written a set of writings.

\paragraph{Deviation Measure}
\label{sec:deviation_measure}
For DDPO and DORPO, we calculate the deviation of each instance ($\delta^{y_i}$). 
Focusing on semantic and style deviations, we calculated deviation for $y_i$ as $\mathbb{E}^N_{j=1, j\neq i} \left[ \bm{d}(y_i, y_j) \right]$ (where $\bm{d}$ is cosine distance after embedding $y_i$ and $Y^x = \{y_i\}^N_{i=1}$).
Note that we used \texttt{jinaai/jina-embeddings-v3} and \texttt{AnnaWegmann/Style-Embedding}, respectively, for semantic and style embedding.
Considering that weights (here, deviations) should be larger than zero and we would want the impact of the total weights to correspond to the size of the dataset per prompt, we transformed deviations to have 1) a minimum of 0 ($\mathrm{min}(\{ \delta^{y_i}\}_{y_i \in Y^x})=0$, unless deviations are all the same) and 2) a sum equal to their count ($\mathrm{sum}(\{ \delta^{y_i}\}_{y_i \in Y^x})=|Y^x|$). We also experimented with mixing semantic and style deviations to consider both diversity types in LLM training. 
We provide details on deviation calculation in Appendix~\ref{app:deviation_transform}.

\paragraph{Conditions} 
As baselines, we examined \textbf{\texttt{SFT}}, \textbf{\texttt{DPO}}, and \textbf{\texttt{ORPO}} as baselines.
We also examined post-trained versions of the base models (trained by base model providers, as \textbf{\texttt{Instruct}}) and four other instruction-tuned models (\texttt{gpt-4o-2024-11-20} as \textbf{\texttt{GPT-4o}}~\citep{gpt4o}, \texttt{o1-2024-12-17} as \textbf{\texttt{o1}}~\citep{o1}, \texttt{claude-3-5-sonnet-20241022} as \textbf{\texttt{Claude-3.5-sonnet}}~\citep{claude3.5sonnet}, and \textbf{\texttt{DeepSeek-R1}}~\citep{deepseekai2025deepseekr1incentivizingreasoningcapability}). 
For \texttt{GPT-4o}, to see whether diverse generation is achievable only with prompting, we evaluated an iterative prompting approach that asks the LLM to generate results far different from previously generated ones (\textbf{\texttt{GPT-4o-iter}}, prompt details are in Appendix~\ref{app:eval_details}). 
We also computed metrics for the original test dataset (\textbf{\texttt{Gold}}).
For DDPO and DORPO, we examined versions that consider only semantic deviation (using \texttt{jinaai/jina-embeddings-v3}, \textbf{\texttt{DDPO-sem}}, \textbf{\texttt{DORPO-sem}}), only style deviation (using \texttt{AnnaWegmann/Style-Embedding}, \textbf{\texttt{DDPO-sty}}, \textbf{\texttt{DORPO-sty}}), and both types of deviations (\textbf{\texttt{DDPO-both}}, \textbf{\texttt{DORPO-both}}).

\paragraph{Training Methods}
We started with the following pre-trained models: \texttt{meta-llama/Llama-3.1-8B}~\citep{grattafiori2024llama3herdmodels} and \texttt{mistralai/Mistral-7B-v0.3}~\citep{jiang2023mistral7b}.\footnote{Accordingly, we used \texttt{meta-llama/Llama-3.1-8B-Instruct} and \texttt{mistralai/Mistral-7B-Instruct-v0.3} for \texttt{Instruct}.}
For SFT, DPO, ORPO, DDPO, and DORPO, we did parameter efficient tuning with LoRA~\citep{hu2022lora}, using a rank of 128 and an alpha of 256. 
This approach optimizes low rank additions to weight matrices and allows efficient tuning of large language models ($\geq$ 7B parameters).

We first trained SFT models as the starting point for DPO training. 
For this, we used a cosine scheduler with a warmup, which went through a half cycle per epoch with a maximum learning rate of 3e-5. We trained with Adam optimizer for a single epoch with a batch size of 2. Evaluation and checkpoint saving were done for every 5000 steps, and we used the models with the lowest evaluation loss. 
Note that all training in this work was done with Accelerate~\citep{accelerate} and DeepSpeed ZeRO-2~\citep{rajbhandari2020zero} in \texttt{bfloat16}, in parallel over six \texttt{NVIDIA H100 SXM} GPUs.

For DPO, DDPO, ORPO, and DORPO, we trained models using the fixed dataset in offline settings. 
For all approaches, we used a linear learning rate with a maximum of 5e-6. Except for DPO and DDPO for \texttt{mistralai/Mistral-7B-v0.3}, the batch size was 2. For DPO and DDPO on \texttt{mistralai/Mistral-7B-v0.3}, we used the batch size of 1 with a gradient accumulation of 2 due to the GPU memory. DPO and DDPO were trained with $\beta$ of 0.1 for three epochs. We trained ORPO and DORPO with $\lambda$ of 0.25 for four epochs, as it starts from the base model, not the SFT model.

\begin{figure}[t]
\begin{center}
    \includegraphics[width=0.99\textwidth]{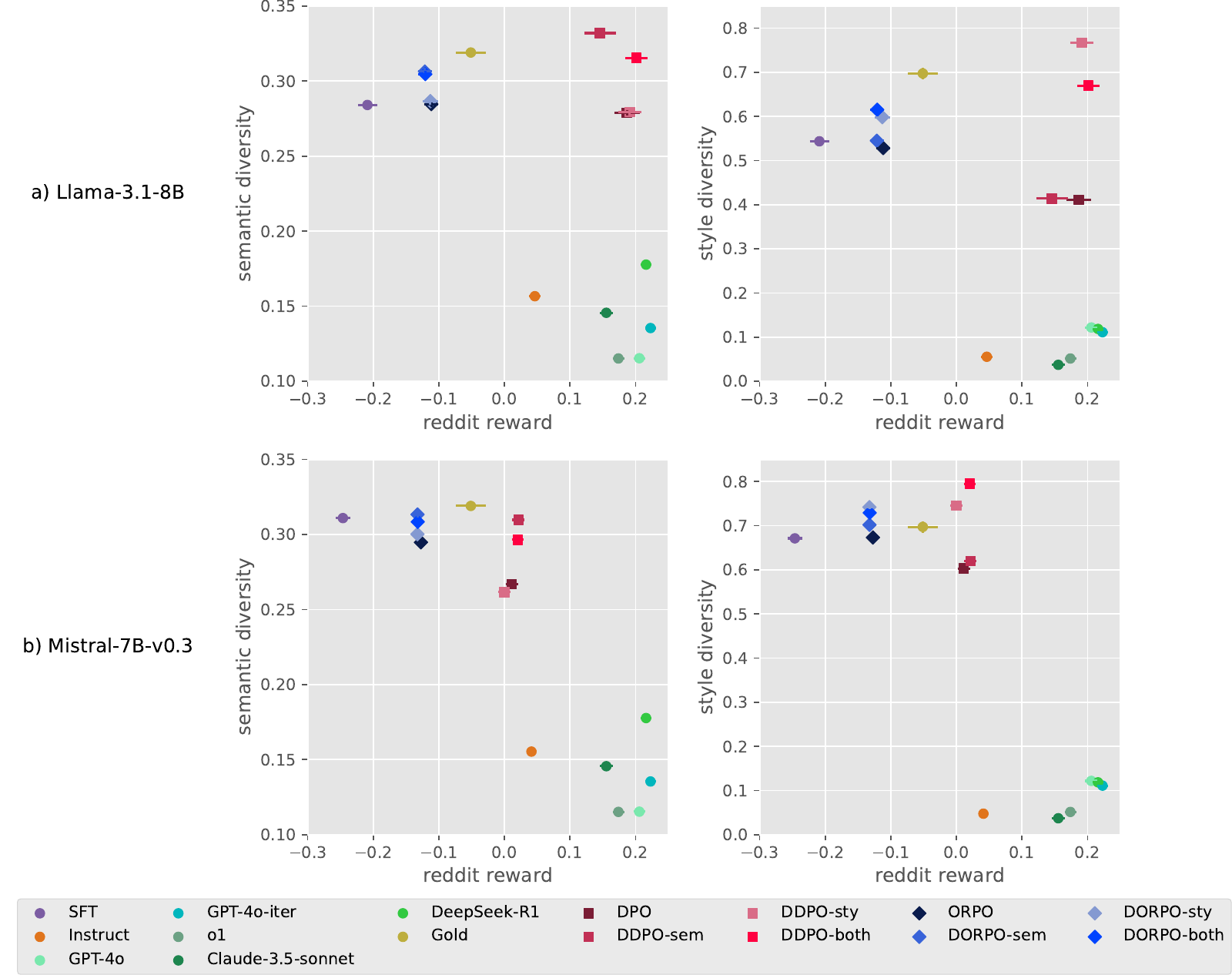}
\end{center}
\caption{Results on writing quality (\texttt{reddit-reward}, $x$ axes) and diversity (\texttt{semantic} or \texttt{style diversity}, $y$ axes). Error bars in this paper indicate 95\% confidence intervals.}
\label{fig:main_result}
\end{figure}

\subsection{Results}
\label{sec:main_results}

\newcommand{\SFTsymbol}[1]{\textcolor[HTML]{7c5da3}{\raisebox{-0.02ex}{\scalebox{1.2}{$\bullet$}}}}
\newcommand{\GPTsymbol}[1]{\textcolor[HTML]{79e8ad}{\raisebox{-0.02ex}{\scalebox{1.2}{$\bullet$}}}}
\newcommand{\GPTitersymbol}[1]{\textcolor[HTML]{00b6bd}{\raisebox{-0.02ex}{\scalebox{1.2}{$\bullet$}}}}
\newcommand{\Oonesymbol}[1]{\textcolor[HTML]{6ca184}{\raisebox{-0.02ex}{\scalebox{1.2}{$\bullet$}}}}
\newcommand{\Claudesymbol}[1]{\textcolor[HTML]{1d854e}{\raisebox{-0.02ex}{\scalebox{1.2}{$\bullet$}}}}
\newcommand{\Deepseeksymbol}[1]{\textcolor[HTML]{32c941}{\raisebox{-0.02ex}{\scalebox{1.2}{$\bullet$}}}}
\newcommand{\Instructsymbol}[1]{\textcolor[HTML]{e0751d}{\raisebox{-0.02ex}{\scalebox{1.2}{$\bullet$}}}}
\newcommand{\Goldsymbol}[1]{\textcolor[HTML]{bdae3e}{\raisebox{-0.02ex}{\scalebox{1.2}{$\bullet$}}}}

\newcommand{\DPOsymbol}[1]{\textcolor[HTML]{7a1d35}{\raisebox{0.14ex}{\rule{0.5em}{0.5em}}}\hspace{0.2ex}}
\newcommand{\DDPOSemsymbol}[1]{\textcolor[HTML]{c23055}{\raisebox{0.14ex}{\rule{0.5em}{0.5em}}}\hspace{0.2ex}}
\newcommand{\DDPOStysymbol}[1]{\textcolor[HTML]{d96c87}{\raisebox{0.14ex}{\rule{0.5em}{0.5em}}}\hspace{0.2ex}}
\newcommand{\DDPOBothsymbol}[1]{\textcolor[HTML]{ff0041}{\raisebox{0.14ex}{\rule{0.5em}{0.5em}}}\hspace{0.2ex}}

\newcommand{\ORPOsymbol}[1]{\textcolor[HTML]{0a1c4d}{\raisebox{-0.03ex}{\rotatebox{45}{\rule{0.5em}{0.5em}}}}\hspace{0.1ex}}
\newcommand{\DORPOSemsymbol}[1]{\textcolor[HTML]{3863d9}{\raisebox{-0.03ex}{\rotatebox{45}{\rule{0.5em}{0.5em}}}}\hspace{0.1ex}}
\newcommand{\DORPOStysymbol}[1]{\textcolor[HTML]{8499d1}{\raisebox{-0.03ex}{\rotatebox{45}{\rule{0.5em}{0.5em}}}}\hspace{0.1ex}}
\newcommand{\DORPOBothsymbol}[1]{\textcolor[HTML]{0044ff}{\raisebox{-0.03ex}{\rotatebox{45}{\rule{0.5em}{0.5em}}}}\hspace{0.1ex}}

Figure~\ref{fig:main_result} shows results on writing quality and diversity. Existing instruction-tuned models (\GPTsymbol{}\texttt{GPT-4o}, \GPTitersymbol{}\texttt{GPT-4o-iter}, \Oonesymbol{}\texttt{o1}, \Claudesymbol{}\texttt{Claude-3.5-sonnet}, \Deepseeksymbol{}\texttt{DeepSeek-R1}, and \Instructsymbol{}\texttt{Instruct})---formed one cluster, where they show high reward but with low diversity. This aligns with previous findings that some post-trained models show low output diversity~\citep{kirk2024understanding, go2023aligning, casper2023open}. 
Human-crafted \Goldsymbol{}\texttt{Gold} showed a lower reward score than these models but the diversity was far higher. 
\SFTsymbol{}\texttt{SFT} models showed the lowest reward scores. Their diversity results were higher than the existing instruction-tuned models, but lower than \Goldsymbol{}\texttt{Gold}. 
Both \DPOsymbol{}\texttt{DPO} and \ORPOsymbol{}\texttt{ORPO} increased \texttt{reddit-reward} compared to \SFTsymbol{}\texttt{SFT} while either maintaining or decreasing diversity. \DPOsymbol{}\texttt{DPO} resulted in higher \texttt{reddit-reward} but often in lower output diversity than \ORPOsymbol{}\texttt{ORPO}. Note that our trained models had higher diversity than existing models (even higher than \GPTitersymbol{}\texttt{GPT-4o-iter}, which used diversity-inducing prompts), which might be due to the diversity of responses for each prompt in our training dataset.

\begin{table}[t]

\caption{Topic sentences of generated examples. Note that ``writing prompt'' is not necessarily a full instruction prompt given to LLMs.} 
\bgroup
\scriptsize
\centering
\def\arraystretch{1.2}
\begin{tabular}{p{0.3\textwidth}p{0.3\textwidth}p{0.3\textwidth}}

\thickhline
\multicolumn{3}{p{0.9\textwidth}}{Writing Prompt: \texttt{"Why are you shaking, my love? You are king now." King can be metaphorical or literal.}}                                                                                                                                                                                                      \\ \thickhline
\multicolumn{1}{c}{\texttt{GPT-4o}}                                                                                              & \multicolumn{1}{c}{\texttt{Llama-3.1-8B DPO}}                                                                                                  & \multicolumn{1}{c}{\texttt{Llama-3.1-8B DDPO-both}}                                                                                                 \\ \thickhline
\texttt{A newly promoted manager struggles with imposter syndrome while his wife offers unwavering support.} & \texttt{A divine ruler must abandon his throne to protect his daughter's reign after revealing a dark secret.} & \texttt{A president grapples with moral conflict over inability to take a life despite leadership expectations.}   \\ \hline
\texttt{A newly crowned king struggles with self-doubt while his beloved offers reassurance.}                & \texttt{A king murders his wife due to a prophecy, while their son watches the tragic sacrifice.}              & \texttt{A couple seeks refuge during a storm, harboring dark secrets and contemplating future redemption.}         \\ \hline
\texttt{A newly crowned king grapples with self-doubt while his beloved offers reassurance.}                 & \texttt{A reluctant new king grapples with the overwhelming burden and fear of leadership.}                    & \texttt{A child's supernatural encounters while diving reveal a dark family tradition of hunting marine spirits.}  \\ \hline
\texttt{A reluctant heir grapples with the weight of unexpected kingship and fear of failure.}               & \texttt{A royal heir struggles with leadership responsibilities while finding strength in romantic love.}      & \texttt{A young bear prince assumes his role as King while facing an arranged marriage in a dark fantasy setting.}

\\\thickhline
\label{tab:examples}
\end{tabular}
\egroup
\end{table}

\textbf{The results show that our approach promotes targeted diversity while minimally decreasing the quality. }
Compared to \DPOsymbol{}\texttt{DPO}, \DDPOSemsymbol{}\texttt{DDPO-sem} and \DDPOStysymbol{}\texttt{DDPO-sty} increased semantic and style diversity, respectively. By mixing two deviation signals, \DDPOBothsymbol{}\texttt{DDPO-both} could facilitate both diversity types. In terms of quality, \texttt{Llama-3.1-8B}'s \DDPOSemsymbol{}\texttt{DDPO-sem} and \texttt{Mistral-7B-v0.3}'s \DDPOStysymbol{}\texttt{DDPO-sty} showed decreased quality compared to \DPOsymbol{}\texttt{DPO} but other \texttt{DDPO} models showed either increased or maintained quality scores. DORPO showed similar patterns: compared to \ORPOsymbol{}\texttt{ORPO}, \DORPOSemsymbol{}\texttt{DORPO-sem}, \DORPOStysymbol{}\texttt{DORPO-sty}, and \DORPOBothsymbol{}\texttt{DORPO-both} increased targeted diversity while minimally decreasing the \texttt{reddit-reward}.
\textbf{Among trained models, \texttt{Llama-3.1-8B}-based \DDPOBothsymbol{}\texttt{DDPO-both} model achieved high scores on both quality and diversity, being almost on-par as the best of baselines and gold data. }
Specifically, it had a \texttt{reddit-reward} only slightly lower than \GPTitersymbol{}\texttt{GPT-4o-iter}, semantic diversity close to \Goldsymbol{}\texttt{Gold}, and style diversity slightly lower than \Goldsymbol{}\texttt{Gold}. For a subset of conditions, we list examples in Table~\ref{tab:examples} (as topic sentences) and Appendix~\ref{app:examples} (as 100-word summaries and 100-word truncations).

\section{Human Evaluation}
\label{sec:human}
We complement the automated evaluation with a human evaluation on a subset of conditions. 
Given two sets of texts from two conditions, evaluators decided 1) which set included the most interesting, highest-quality writing and 2) which set was more diverse. If a decision was difficult to make, they could answer with a ``Hard to decide'' option.
Each set included four creative writings generated by the model for that condition. Note that we provided summarized versions of writings as doing the task with eight lengthy creative writings can be cognitively overloading. Hence, with human evaluation, we could measure only semantic diversity but not style diversity.
Five of this paper's authors served as evaluators, being blind to the conditions. Three evaluated each instance, and we aggregated their annotations with majority voting.
With this approach, we evaluated \texttt{DPO} vs. \texttt{DDPO-both} and \texttt{GPT-4o} vs. \texttt{DDPO-both}. Over other approaches for existing instruction-tuned models, we choose to compare with \texttt{GPT-4o} due to 1) its popularity in LLM research and products and 2) the simplicity in prompting. 
Refer to Appendix~\ref{app:humaneval} for more details.

\begin{table}[t]
\caption{Win rates(\%) from human evaluation. Significant differences are bolded.}
\small
\centering
\def\arraystretch{1.1}

\begin{tabular}{l|lll|lll}
\thickhline
                                & \texttt{DDPO-both}  & vs. & \texttt{GPT-4o} & \texttt{DDPO-both} & vs. & \texttt{DPO}  \\ \hline
Has the highest-quality story & \textbf{68\%}    &  & \textbf{24\%}   & 50\%    &  & 34\% \\
More diverse                    & \textbf{100\%}   &  & \textbf{0\%}    & \textbf{62\%}    &  & \textbf{26\%} \\ \thickhline
\end{tabular}
\label{tab:humaneval}
\end{table}

\paragraph{Results}
When compared to \texttt{GPT-4o}, evaluators chose \texttt{DDPO-both}'s sets more frequently as those that have the highest-quality story (Table~\ref{tab:humaneval}). The ratio difference was significant with Two Proportion Z-Test ($p<0.001$).
When compared to \texttt{DPO}, \texttt{DDPO-both} was chosen more frequently, but the difference in ratio was not significant ($p>0.1$).
Regarding diversity, evaluators chose \texttt{DDPO-both}'s sets more frequently as more diverse sets, compared to both \texttt{GPT-4o} and \texttt{DPO}. The ratio differences were significant for both comparisons ($p<0.001$ for both).
Evaluator agreement (Krippendorff's alpha) was 0.31 and 0.45, respectively, for quality and diversity selections, indicating fair agreement. We count in the evaluators' indecision when calculating agreement values.
When calculate agreement scores separately for vs. \texttt{GPT-4o} and vs. \texttt{DPO}, quality agreement scores were 0.37 and 0.27, respectively, and diversity agreement scores were 0.95 and 0.12, respectively. This reflects that smaller diversity differences in vs. \texttt{DPO} were more difficult to discern than for those in vs. \texttt{GPT-4o}.
\section{Ablation and Comparison to DivPO}

While Section~\ref{sec:main_results} and~\ref{sec:human} show that diversification approaches work for the examined dataset, we were curious if the approaches would still work when the size of the dataset, specifically, the number of responses per prompt, is small. This is an important question, as crafting a dataset with many instances per prompt can be expensive. Hence, we conducted an ablation study, evaluating the performance of trained models when we varied the maximum number of responses per prompt. Specifically, we compared \texttt{DDPO-both} to \texttt{DPO} for different numbers of responses, from four up to the maximum provided in the dataset. 

Here, we also compare our (ablated) approaches to a recent tuning approach for facilitating output diversity, DivPO~\citep{lanchantin2025diversepreferenceoptimization}. 
When using our dataset in full, we cannot apply DivPO, as it requires filtering instances based on their quality and diversity. However, when limiting the maximum number of responses per prompt ($n_\mathrm{max}$), we need to sample a subset of instances and we can apply DivPO's filtering approach (detailed in Appendix~\ref{app:divpo}).\footnote{Note that, practically, if we want to train models with the same number of instances, DivPO requires more data than our approaches due to filtering.}
We considered both semantic and style diversity signals when applying DivPO.

\paragraph{Results}
\begin{figure}[t]
\begin{center}
    \includegraphics[width=\textwidth]{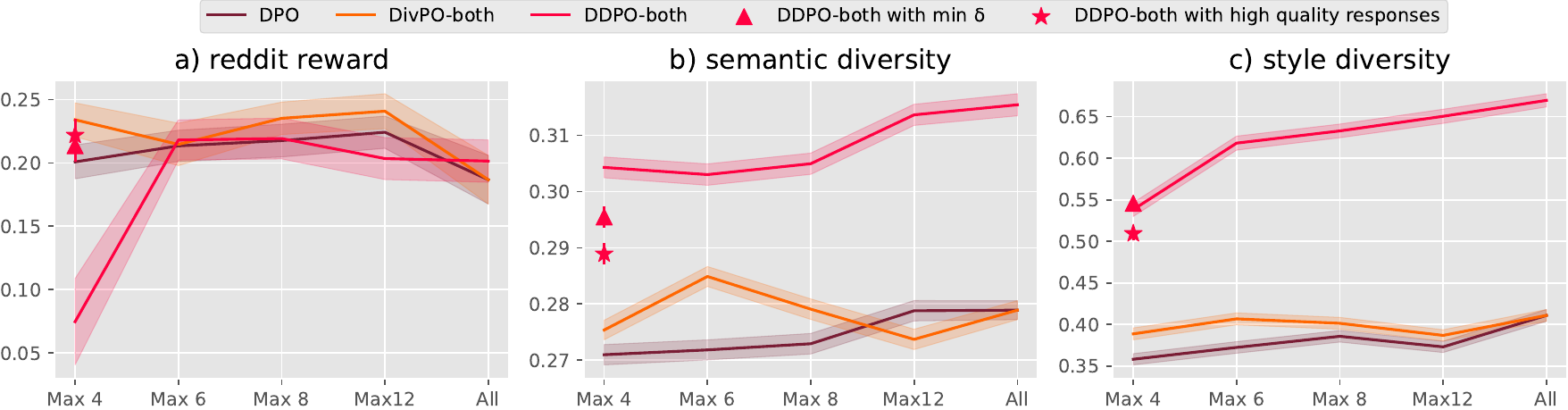}
\end{center}
\caption{Ablation results by varying the maximum number of responses per prompt. When the maximum number of responses is four, we also experimented with 1) setting a minimum $\delta$ and 2) using high-quality responses.}
\label{fig:ablation}
\end{figure}

\newcommand{\DPOabsymbol}[1]{\textcolor[HTML]{7a1d35}{\raisebox{0.25em}{\rule{0.5em}{0.12em}}}\hspace{0.2ex}}
\newcommand{\DDPOBothabsymbol}[1]{\textcolor[HTML]{ff0041}{\raisebox{0.25em}{\rule{0.5em}{0.12em}}}\hspace{0.2ex}}
\newcommand{\DivPOabsymbol}[1]{\textcolor[HTML]{ff6600}{\raisebox{0.25em}{\rule{0.5em}{0.12em}}}\hspace{0.2ex}}
\newcommand{\DDPOBothminabsymbol}[1]{\textcolor[HTML]{ff0041}{$\blacktriangle$}}

\newcommand{\DDPOBothhighabsymbol}[1]{\textcolor[HTML]{ff0041}{$\bigstar$}}

Figure~\ref{fig:ablation} shows the results, where \textbf{except for when the maximum number of responses is four, \DDPOBothabsymbol{}\texttt{DDPO-both} had similar or only slightly lower mean \texttt{reddit-reward}s to \DPOabsymbol{}\texttt{DPO} and \DivPOabsymbol{}\texttt{DivPO} while showing higher diversity than them}.
\DivPOabsymbol{}\texttt{DivPO} showed diversity scores slightly higher than \DPOabsymbol{}\texttt{DPO} except for the \texttt{semantic diversity} when the maximum number of instances per prompt was 12. Still, \DivPOabsymbol{}\texttt{DivPO}'s diversity scores were lower than those of \DDPOBothabsymbol{}\texttt{DDPO-both}.
Note that \DivPOabsymbol{}\texttt{DivPO} tends to show slightly higher \texttt{reddit-reward}s than other approaches, except for when the maximum number of instances per prompt was six or no filtering was used. 
This might be because \DivPOabsymbol{}\texttt{DivPO} first sifts through the dataset to get the highest-quality winning instances and lowest-quality losing instances. 
\DPOabsymbol{}\texttt{DPO} and \DDPOBothabsymbol{}\texttt{DDPO-both} randomly sampled instances without such filtering.
Overall, the results demonstrate that our approach is more effective in promoting diversity while not hurting the quality much except when there are too few responses per prompt. 

For the result when the maximum number of responses per prompt is four, we hypothesized that the decrease in \texttt{reddit-reward} might be because there can be cases where $\delta^w$ is zero (in Equation~\ref{eq:ddpo}) while such an issue does not happen when we do not scale pairs with deviation.
With few instances per prompt, the ratio of pairs having zero $\delta^w$ would be higher, which could have impacted the performance on \texttt{reddit-reward} more.

We examined whether this issue could be addressed by forcing a non-zero minimum $\delta^w$ value. That is, in \DDPOBothminabsymbol{}\texttt{DDPO-both with min $\delta$}, we replaced $\delta^w$ smaller than a threshold value (0.1, in our examination) to the threshold value. The result showed that setting the minimum $\delta^w$ helped with increasing \texttt{reddit-reward} to the level of \DPOabsymbol{}\texttt{DPO}'s while decreasing the amount of boost in the diversification approach, specifically in semantic diversity. However, its diversity scores were still higher than \DPOabsymbol{}\texttt{DPO} and \DivPOabsymbol{}\texttt{DivPO}.

We were also curious if this issue could be alleviated if we prepared higher-quality winning responses. Hence, we also examined how the performance changes when we had the highest quality winning responses when sampling at most four responses per prompt (\DDPOBothhighabsymbol{}\texttt{DDPO-both with high-quality responses}). We found that this approach resulted in a mean \texttt{reddit-reward} higher than \DPOabsymbol{}\texttt{DPO} and only slightly lower than \DivPOabsymbol{}\texttt{DivPO}.
Both types of diversity decreased compared to \DDPOBothabsymbol{}\texttt{DDPO-both}, but they were still largely higher than those of \DPOabsymbol{}\texttt{DPO} and \DivPOabsymbol{}\texttt{DivPO}. \textbf{Overall, by either tweaking $\delta^w$ or preparing high-quality data, we could address DDPO's generation quality issue when the number of responses per prompt is small but at the cost of diversity scores. Despite this diversity cost, DDPO still obtains higher diversity than DPO and DivPO. }
\section{Concluding Discussion}

In this work, in a creative writing context, we introduce extended versions of DPO and ORPO that facilitate diversity while maintaining generation quality. The core idea behind our extensions is factoring in the deviation of each winning instance to loss terms. With these approaches, we achieved a model that has quality on par with existing state-of-the-art models and diversity similar to the original human-crafted datasets. We also demonstrate that our approaches could be robust to variation in dataset size while outperforming an existing diversification post-training approach, DivPO. Note that while DivPO requires more data instances than actually used for training (as it filters data), our approach fully uses a given dataset---which would be valuable for data-scarce settings. Overall, our approach emphasizes that to facilitate generation diversity, it is important to balance learning from both frequent and rare high-quality training instances.

We demonstrated the benefits of our approach in our setting, but validating the approach in other settings would be important. Specifically, many instruction-tuned models with low output diversity were trained with online approaches, and future work would need to investigate whether approaches similar to ours can alleviate such issues in online training settings. Moreover, future work would need to examine diversifying tuning approaches in tasks other than creative writing. In addition, as we demonstrated, there can be multiple ways to configure the deviation term, and further exploring these can be future work. As we used deviation terms with winning-losing response pairs, another future work could explore how we can adopt deviation terms in tuning approaches that do not use winning-losing pairs, such as those that use numerical rewards.

\section*{Acknowledgments}
\makeatletter
\@ifpackagewith{colm2025_conference}{submission}{Anonymized}{We want to thank Midjourney for supporting this work.}
\makeatother

\bibliography{colm2025_conference}

\begin{thebibliography}{48}
\providecommand{\natexlab}[1]{#1}
\providecommand{\url}[1]{\texttt{#1}}
\expandafter\ifx\csname urlstyle\endcsname\relax
  \providecommand{\doi}[1]{doi: #1}\else
  \providecommand{\doi}{doi: \begingroup \urlstyle{rm}\Url}\fi

\bibitem[Anderson et~al.(2024)Anderson, Shah, and Kreminski]{anderson2024homogenization}
Barrett~R Anderson, Jash~Hemant Shah, and Max Kreminski.
\newblock Homogenization effects of large language models on human creative ideation.
\newblock In \emph{Proceedings of the 16th Conference on Creativity \& Cognition}, C\&C '24, pp.\  413–425, New York, NY, USA, 2024. Association for Computing Machinery.
\newblock ISBN 9798400704857.
\newblock \doi{10.1145/3635636.3656204}.
\newblock URL \url{https://doi.org/10.1145/3635636.3656204}.

\bibitem[Anthropic(2024)]{claude3.5sonnet}
Anthropic.
\newblock Claude 3.5 sonnet, 2024.
\newblock URL \url{https://www.anthropic.com/news/claude-3-5-sonnet}.
\newblock Accessed: February, 2025.

\bibitem[Bradley et~al.(2024)Bradley, Dai, Teufel, Zhang, Oostermeijer, Bellagente, Clune, Stanley, Schott, and Lehman]{bradley2024qualitydiversity}
Herbie Bradley, Andrew Dai, Hannah~Benita Teufel, Jenny Zhang, Koen Oostermeijer, Marco Bellagente, Jeff Clune, Kenneth Stanley, Gregory Schott, and Joel Lehman.
\newblock Quality-diversity through {AI} feedback.
\newblock In \emph{The Twelfth International Conference on Learning Representations}, 2024.
\newblock URL \url{https://openreview.net/forum?id=owokKCrGYr}.

\bibitem[Casper et~al.(2023)Casper, Davies, Shi, Gilbert, Scheurer, Rando, Freedman, Korbak, Lindner, Freire, Wang, Marks, Segerie, Carroll, Peng, Christoffersen, Damani, Slocum, Anwar, Siththaranjan, Nadeau, Michaud, Pfau, Krasheninnikov, Chen, Langosco, Hase, Biyik, Dragan, Krueger, Sadigh, and Hadfield-Menell]{casper2023open}
Stephen Casper, Xander Davies, Claudia Shi, Thomas~Krendl Gilbert, J{\'e}r{\'e}my Scheurer, Javier Rando, Rachel Freedman, Tomek Korbak, David Lindner, Pedro Freire, Tony~Tong Wang, Samuel Marks, Charbel-Raphael Segerie, Micah Carroll, Andi Peng, Phillip~J.K. Christoffersen, Mehul Damani, Stewart Slocum, Usman Anwar, Anand Siththaranjan, Max Nadeau, Eric~J Michaud, Jacob Pfau, Dmitrii Krasheninnikov, Xin Chen, Lauro Langosco, Peter Hase, Erdem Biyik, Anca Dragan, David Krueger, Dorsa Sadigh, and Dylan Hadfield-Menell.
\newblock Open problems and fundamental limitations of reinforcement learning from human feedback.
\newblock \emph{Transactions on Machine Learning Research}, 2023.
\newblock ISSN 2835-8856.
\newblock URL \url{https://openreview.net/forum?id=bx24KpJ4Eb}.
\newblock Survey Certification, Featured Certification.

\bibitem[Chen et~al.(2025)Chen, Song, Zheng, Jing, Hansen, and Sun]{chen2025understandingdesignfixationgenerative}
Liuqing Chen, Yaxuan Song, Chunyuan Zheng, Qianzhi Jing, Preben Hansen, and Lingyun Sun.
\newblock Understanding design fixation in generative ai, 2025.
\newblock URL \url{https://arxiv.org/abs/2502.05870}.

\bibitem[Chung et~al.(2023)Chung, Kamar, and Amershi]{chung2023increasing}
John Chung, Ece Kamar, and Saleema Amershi.
\newblock Increasing diversity while maintaining accuracy: Text data generation with large language models and human interventions.
\newblock In Anna Rogers, Jordan Boyd-Graber, and Naoaki Okazaki (eds.), \emph{Proceedings of the 61st Annual Meeting of the Association for Computational Linguistics (Volume 1: Long Papers)}, pp.\  575--593, Toronto, Canada, July 2023. Association for Computational Linguistics.
\newblock \doi{10.18653/v1/2023.acl-long.34}.
\newblock URL \url{https://aclanthology.org/2023.acl-long.34/}.

\bibitem[Cox et~al.(2021)Cox, Wang, Abdul, von~der Weth, and Y.~Lim]{cox2021directeddiversity}
Samuel~Rhys Cox, Yunlong Wang, Ashraf Abdul, Christian von~der Weth, and Brian Y.~Lim.
\newblock Directed diversity: Leveraging language embedding distances for collective creativity in crowd ideation.
\newblock In \emph{Proceedings of the 2021 CHI Conference on Human Factors in Computing Systems}, CHI '21, New York, NY, USA, 2021. Association for Computing Machinery.
\newblock ISBN 9781450380966.
\newblock \doi{10.1145/3411764.3445782}.
\newblock URL \url{https://doi.org/10.1145/3411764.3445782}.

\bibitem[DeepSeek-AI et~al.(2025)DeepSeek-AI, Guo, Yang, Zhang, Song, Zhang, Xu, Zhu, Ma, Wang, Bi, and et~al.]{deepseekai2025deepseekr1incentivizingreasoningcapability}
DeepSeek-AI, Daya Guo, Dejian Yang, Haowei Zhang, Junxiao Song, Ruoyu Zhang, Runxin Xu, Qihao Zhu, Shirong Ma, Peiyi Wang, Xiao Bi, and et~al.
\newblock Deepseek-r1: Incentivizing reasoning capability in llms via reinforcement learning, 2025.
\newblock URL \url{https://arxiv.org/abs/2501.12948}.

\bibitem[Fan et~al.(2018)Fan, Lewis, and Dauphin]{fan2018hierarchical}
Angela Fan, Mike Lewis, and Yann Dauphin.
\newblock Hierarchical neural story generation.
\newblock In Iryna Gurevych and Yusuke Miyao (eds.), \emph{Proceedings of the 56th Annual Meeting of the Association for Computational Linguistics (Volume 1: Long Papers)}, pp.\  889--898, Melbourne, Australia, July 2018. Association for Computational Linguistics.
\newblock \doi{10.18653/v1/P18-1082}.
\newblock URL \url{https://aclanthology.org/P18-1082/}.

\bibitem[Flower \& Hayes(1981)Flower and Hayes]{flower1981cognitive}
Linda Flower and John~R. Hayes.
\newblock A cognitive process theory of writing.
\newblock \emph{College Composition and Communication}, 32\penalty0 (4):\penalty0 365--387, 1981.
\newblock ISSN 0010096X.
\newblock URL \url{http://www.jstor.org/stable/356600}.

\bibitem[Go et~al.(2023)Go, Korbak, Kruszewski, Rozen, Ryu, and Dymetman]{go2023aligning}
Dongyoung Go, Tomasz Korbak, Germ\'{a}n Kruszewski, Jos Rozen, Nahyeon Ryu, and Marc Dymetman.
\newblock Aligning language models with preferences through f-divergence minimization.
\newblock In \emph{Proceedings of the 40th International Conference on Machine Learning}, ICML'23. JMLR.org, 2023.

\bibitem[Grattafiori et~al.(2024)Grattafiori, Dubey, Jauhri, Pandey, Kadian, Al-Dahle, Letman, Mathur, Schelten, Vaughan, and et~al.]{grattafiori2024llama3herdmodels}
Aaron Grattafiori, Abhimanyu Dubey, Abhinav Jauhri, Abhinav Pandey, Abhishek Kadian, Ahmad Al-Dahle, Aiesha Letman, Akhil Mathur, Alan Schelten, Alex Vaughan, and et~al.
\newblock The llama 3 herd of models, 2024.
\newblock URL \url{https://arxiv.org/abs/2407.21783}.

\bibitem[Gugger et~al.(2022)Gugger, Debut, Wolf, Schmid, Mueller, Mangrulkar, Sun, and Bossan]{accelerate}
Sylvain Gugger, Lysandre Debut, Thomas Wolf, Philipp Schmid, Zachary Mueller, Sourab Mangrulkar, Marc Sun, and Benjamin Bossan.
\newblock Accelerate: Training and inference at scale made simple, efficient and adaptable.
\newblock \url{https://github.com/huggingface/accelerate}, 2022.

\bibitem[Guilford(1957)]{guilford1957creative}
Joy~P Guilford.
\newblock Creative abilities in the arts.
\newblock \emph{Psychological review}, 64\penalty0 (2):\penalty0 110, 1957.

\bibitem[Hayati et~al.(2024)Hayati, Lee, Rajagopal, and Kang]{hayati2024far}
Shirley~Anugrah Hayati, Minhwa Lee, Dheeraj Rajagopal, and Dongyeop Kang.
\newblock How far can we extract diverse perspectives from large language models?
\newblock In Yaser Al-Onaizan, Mohit Bansal, and Yun-Nung Chen (eds.), \emph{Proceedings of the 2024 Conference on Empirical Methods in Natural Language Processing}, pp.\  5336--5366, Miami, Florida, USA, November 2024. Association for Computational Linguistics.
\newblock \doi{10.18653/v1/2024.emnlp-main.306}.
\newblock URL \url{https://aclanthology.org/2024.emnlp-main.306/}.

\bibitem[Holtzman et~al.(2020)Holtzman, Buys, Du, Forbes, and Choi]{Holtzman2020The}
Ari Holtzman, Jan Buys, Li~Du, Maxwell Forbes, and Yejin Choi.
\newblock The curious case of neural text degeneration.
\newblock In \emph{International Conference on Learning Representations}, 2020.
\newblock URL \url{https://openreview.net/forum?id=rygGQyrFvH}.

\bibitem[Hong et~al.(2024)Hong, Lee, and Thorne]{hong2024orpomonolithicpreferenceoptimization}
Jiwoo Hong, Noah Lee, and James Thorne.
\newblock Orpo: Monolithic preference optimization without reference model, 2024.
\newblock URL \url{https://arxiv.org/abs/2403.07691}.

\bibitem[Hu et~al.(2022)Hu, yelong shen, Wallis, Allen-Zhu, Li, Wang, Wang, and Chen]{hu2022lora}
Edward~J Hu, yelong shen, Phillip Wallis, Zeyuan Allen-Zhu, Yuanzhi Li, Shean Wang, Lu~Wang, and Weizhu Chen.
\newblock Lo{RA}: Low-rank adaptation of large language models.
\newblock In \emph{International Conference on Learning Representations}, 2022.
\newblock URL \url{https://openreview.net/forum?id=nZeVKeeFYf9}.

\bibitem[Ippolito et~al.(2019)Ippolito, Kriz, Sedoc, Kustikova, and Callison-Burch]{ippolito2019comparison}
Daphne Ippolito, Reno Kriz, Jo{\~a}o Sedoc, Maria Kustikova, and Chris Callison-Burch.
\newblock Comparison of diverse decoding methods from conditional language models.
\newblock In Anna Korhonen, David Traum, and Llu{\'i}s M{\`a}rquez (eds.), \emph{Proceedings of the 57th Annual Meeting of the Association for Computational Linguistics}, pp.\  3752--3762, Florence, Italy, July 2019. Association for Computational Linguistics.
\newblock \doi{10.18653/v1/P19-1365}.
\newblock URL \url{https://aclanthology.org/P19-1365/}.

\bibitem[Jakesch et~al.(2023)Jakesch, Bhat, Buschek, Zalmanson, and Naaman]{jakesch2023cowriting}
Maurice Jakesch, Advait Bhat, Daniel Buschek, Lior Zalmanson, and Mor Naaman.
\newblock Co-writing with opinionated language models affects users’ views.
\newblock In \emph{Proceedings of the 2023 CHI Conference on Human Factors in Computing Systems}, CHI '23, New York, NY, USA, 2023. Association for Computing Machinery.
\newblock ISBN 9781450394215.
\newblock \doi{10.1145/3544548.3581196}.
\newblock URL \url{https://doi.org/10.1145/3544548.3581196}.

\bibitem[Jiang et~al.(2023)Jiang, Sablayrolles, Mensch, Bamford, Chaplot, de~las Casas, Bressand, Lengyel, Lample, Saulnier, Lavaud, Lachaux, Stock, Scao, Lavril, Wang, Lacroix, and Sayed]{jiang2023mistral7b}
Albert~Q. Jiang, Alexandre Sablayrolles, Arthur Mensch, Chris Bamford, Devendra~Singh Chaplot, Diego de~las Casas, Florian Bressand, Gianna Lengyel, Guillaume Lample, Lucile Saulnier, Lélio~Renard Lavaud, Marie-Anne Lachaux, Pierre Stock, Teven~Le Scao, Thibaut Lavril, Thomas Wang, Timothée Lacroix, and William~El Sayed.
\newblock Mistral 7b, 2023.
\newblock URL \url{https://arxiv.org/abs/2310.06825}.

\bibitem[Kirk et~al.(2024)Kirk, Mediratta, Nalmpantis, Luketina, Hambro, Grefenstette, and Raileanu]{kirk2024understanding}
Robert Kirk, Ishita Mediratta, Christoforos Nalmpantis, Jelena Luketina, Eric Hambro, Edward Grefenstette, and Roberta Raileanu.
\newblock Understanding the effects of {RLHF} on {LLM} generalisation and diversity.
\newblock In \emph{The Twelfth International Conference on Learning Representations}, 2024.
\newblock URL \url{https://openreview.net/forum?id=PXD3FAVHJT}.

\bibitem[Lahoti et~al.(2023)Lahoti, Blumm, Ma, Kotikalapudi, Potluri, Tan, Srinivasan, Packer, Beirami, Beutel, and Chen]{lahoti2023improving}
Preethi Lahoti, Nicholas Blumm, Xiao Ma, Raghavendra Kotikalapudi, Sahitya Potluri, Qijun Tan, Hansa Srinivasan, Ben Packer, Ahmad Beirami, Alex Beutel, and Jilin Chen.
\newblock Improving diversity of demographic representation in large language models via collective-critiques and self-voting.
\newblock In Houda Bouamor, Juan Pino, and Kalika Bali (eds.), \emph{Proceedings of the 2023 Conference on Empirical Methods in Natural Language Processing}, pp.\  10383--10405, Singapore, December 2023. Association for Computational Linguistics.
\newblock \doi{10.18653/v1/2023.emnlp-main.643}.
\newblock URL \url{https://aclanthology.org/2023.emnlp-main.643/}.

\bibitem[Lanchantin et~al.(2025)Lanchantin, Chen, Dhuliawala, Yu, Weston, Sukhbaatar, and Kulikov]{lanchantin2025diversepreferenceoptimization}
Jack Lanchantin, Angelica Chen, Shehzaad Dhuliawala, Ping Yu, Jason Weston, Sainbayar Sukhbaatar, and Ilia Kulikov.
\newblock Diverse preference optimization, 2025.
\newblock URL \url{https://arxiv.org/abs/2501.18101}.

\bibitem[Li et~al.(2016)Li, Monroe, and Jurafsky]{li2016simplefastdiversedecoding}
Jiwei Li, Will Monroe, and Dan Jurafsky.
\newblock A simple, fast diverse decoding algorithm for neural generation, 2016.
\newblock URL \url{https://arxiv.org/abs/1611.08562}.

\bibitem[Minh et~al.(2025)Minh, Baker, Neo, Roush, Kirsch, and Shwartz-Ziv]{minh2025turning}
Nguyen~Nhat Minh, Andrew Baker, Clement Neo, Allen~G Roush, Andreas Kirsch, and Ravid Shwartz-Ziv.
\newblock Turning up the heat: Min-p sampling for creative and coherent {LLM} outputs.
\newblock In \emph{The Thirteenth International Conference on Learning Representations}, 2025.
\newblock URL \url{https://openreview.net/forum?id=FBkpCyujtS}.

\bibitem[Narayanan~Venkit et~al.(2023)Narayanan~Venkit, Gautam, Panchanadikar, Huang, and Wilson]{venkit2023unmasking}
Pranav Narayanan~Venkit, Sanjana Gautam, Ruchi Panchanadikar, Ting-Hao Huang, and Shomir Wilson.
\newblock Unmasking nationality bias: A study of human perception of nationalities in ai-generated articles.
\newblock In \emph{Proceedings of the 2023 AAAI/ACM Conference on AI, Ethics, and Society}, AIES '23, pp.\  554–565, New York, NY, USA, 2023. Association for Computing Machinery.
\newblock ISBN 9798400702310.
\newblock \doi{10.1145/3600211.3604667}.
\newblock URL \url{https://doi.org/10.1145/3600211.3604667}.

\bibitem[OpenAI(2024{\natexlab{a}})]{gpt4o}
OpenAI.
\newblock Hello gpt-4o, 2024{\natexlab{a}}.
\newblock URL \url{https://openai.com/index/hello-gpt-4o/}.
\newblock Accessed: February, 2025.

\bibitem[OpenAI(2024{\natexlab{b}})]{o1}
OpenAI.
\newblock Introducing openai o1, 2024{\natexlab{b}}.
\newblock URL \url{https://openai.com/o1/}.
\newblock Accessed: February, 2025.

\bibitem[Ouyang et~al.(2022)Ouyang, Wu, Jiang, Almeida, Wainwright, Mishkin, Zhang, Agarwal, Slama, Ray, Schulman, Hilton, Kelton, Miller, Simens, Askell, Welinder, Christiano, Leike, and Lowe]{ouyang2022training}
Long Ouyang, Jeffrey Wu, Xu~Jiang, Diogo Almeida, Carroll Wainwright, Pamela Mishkin, Chong Zhang, Sandhini Agarwal, Katarina Slama, Alex Ray, John Schulman, Jacob Hilton, Fraser Kelton, Luke Miller, Maddie Simens, Amanda Askell, Peter Welinder, Paul~F Christiano, Jan Leike, and Ryan Lowe.
\newblock Training language models to follow instructions with human feedback.
\newblock In S.~Koyejo, S.~Mohamed, A.~Agarwal, D.~Belgrave, K.~Cho, and A.~Oh (eds.), \emph{Advances in Neural Information Processing Systems}, volume~35, pp.\  27730--27744. Curran Associates, Inc., 2022.
\newblock URL \url{https://proceedings.neurips.cc/paper_files/paper/2022/file/b1efde53be364a73914f58805a001731-Paper-Conference.pdf}.

\bibitem[Padmakumar \& He(2024)Padmakumar and He]{padmakumar2024does}
Vishakh Padmakumar and He~He.
\newblock Does writing with language models reduce content diversity?
\newblock In \emph{The Twelfth International Conference on Learning Representations}, 2024.
\newblock URL \url{https://openreview.net/forum?id=Feiz5HtCD0}.

\bibitem[Pei et~al.(2022)Pei, Ananthasubramaniam, Wang, Zhou, Dedeloudis, Sargent, and Jurgens]{pei2022potato}
Jiaxin Pei, Aparna Ananthasubramaniam, Xingyao Wang, Naitian Zhou, Apostolos Dedeloudis, Jackson Sargent, and David Jurgens.
\newblock Potato: The portable text annotation tool.
\newblock In \emph{Proceedings of the 2022 Conference on Empirical Methods in Natural Language Processing: System Demonstrations}, 2022.

\bibitem[Rafailov et~al.(2023)Rafailov, Sharma, Mitchell, Manning, Ermon, and Finn]{rafailov2023direct}
Rafael Rafailov, Archit Sharma, Eric Mitchell, Christopher~D Manning, Stefano Ermon, and Chelsea Finn.
\newblock Direct preference optimization: Your language model is secretly a reward model.
\newblock In A.~Oh, T.~Naumann, A.~Globerson, K.~Saenko, M.~Hardt, and S.~Levine (eds.), \emph{Advances in Neural Information Processing Systems}, volume~36, pp.\  53728--53741. Curran Associates, Inc., 2023.
\newblock URL \url{https://proceedings.neurips.cc/paper_files/paper/2023/file/a85b405ed65c6477a4fe8302b5e06ce7-Paper-Conference.pdf}.

\bibitem[Rajbhandari et~al.(2020)Rajbhandari, Rasley, Ruwase, and He]{rajbhandari2020zero}
Samyam Rajbhandari, Jeff Rasley, Olatunji Ruwase, and Yuxiong He.
\newblock Zero: memory optimizations toward training trillion parameter models.
\newblock In \emph{Proceedings of the International Conference for High Performance Computing, Networking, Storage and Analysis}, SC '20. IEEE Press, 2020.
\newblock ISBN 9781728199986.

\bibitem[Runco \& Acar(2012)Runco and Acar]{runco2012divergent}
Mark~A. Runco and Selcuk Acar.
\newblock Divergent thinking as an indicator of creative potential.
\newblock \emph{Creativity Research Journal}, 24\penalty0 (1):\penalty0 66--75, 2012.
\newblock \doi{10.1080/10400419.2012.652929}.

\bibitem[Shaib et~al.(2024)Shaib, Barrow, Sun, Siu, Wallace, and Nenkova]{shaib2024standardizingmeasurementtextdiversity}
Chantal Shaib, Joe Barrow, Jiuding Sun, Alexa~F. Siu, Byron~C. Wallace, and Ani Nenkova.
\newblock Standardizing the measurement of text diversity: A tool and a comparative analysis of scores, 2024.
\newblock URL \url{https://arxiv.org/abs/2403.00553}.

\bibitem[Sturua et~al.(2024)Sturua, Mohr, Akram, Günther, Wang, Krimmel, Wang, Mastrapas, Koukounas, Koukounas, Wang, and Xiao]{sturua2024jinaembeddingsv3multilingualembeddingstask}
Saba Sturua, Isabelle Mohr, Mohammad~Kalim Akram, Michael Günther, Bo~Wang, Markus Krimmel, Feng Wang, Georgios Mastrapas, Andreas Koukounas, Andreas Koukounas, Nan Wang, and Han Xiao.
\newblock jina-embeddings-v3: Multilingual embeddings with task lora, 2024.
\newblock URL \url{https://arxiv.org/abs/2409.10173}.

\bibitem[Suh et~al.(2024)Suh, Chen, Min, Li, and Xia]{suh2024luminate}
Sangho Suh, Meng Chen, Bryan Min, Toby Jia-Jun Li, and Haijun Xia.
\newblock Luminate: Structured generation and exploration of design space with large language models for human-ai co-creation.
\newblock In \emph{Proceedings of the 2024 CHI Conference on Human Factors in Computing Systems}, CHI '24, New York, NY, USA, 2024. Association for Computing Machinery.
\newblock ISBN 9798400703300.
\newblock \doi{10.1145/3613904.3642400}.
\newblock URL \url{https://doi.org/10.1145/3613904.3642400}.

\bibitem[Tevet \& Berant(2021)Tevet and Berant]{tevet2021evaluating}
Guy Tevet and Jonathan Berant.
\newblock Evaluating the evaluation of diversity in natural language generation.
\newblock In Paola Merlo, Jorg Tiedemann, and Reut Tsarfaty (eds.), \emph{Proceedings of the 16th Conference of the European Chapter of the Association for Computational Linguistics: Main Volume}, pp.\  326--346, Online, April 2021. Association for Computational Linguistics.
\newblock \doi{10.18653/v1/2021.eacl-main.25}.
\newblock URL \url{https://aclanthology.org/2021.eacl-main.25/}.

\bibitem[Venkit et~al.(2024)Venkit, Laban, Zhou, Mao, and Wu]{venkit2024searchenginesaiera}
Pranav~Narayanan Venkit, Philippe Laban, Yilun Zhou, Yixin Mao, and Chien-Sheng Wu.
\newblock Search engines in an ai era: The false promise of factual and verifiable source-cited responses, 2024.
\newblock URL \url{https://arxiv.org/abs/2410.22349}.

\bibitem[Vijayakumar et~al.(2017)Vijayakumar, Cogswell, Selvaraju, Sun, Lee, Crandall, and Batra]{vijayakumar2017diverse}
Ashwin~K Vijayakumar, Michael Cogswell, Ramprasaath~R. Selvaraju, Qing Sun, Stefan Lee, David Crandall, and Dhruv Batra.
\newblock Diverse beam search: Decoding diverse solutions from neural sequence models, 2017.
\newblock URL \url{https://openreview.net/forum?id=HJV1zP5xg}.

\bibitem[Wang \& Kreminski(2024)Wang and Kreminski]{wang2024guidingdiversifyingllmbasedstory}
Phoebe~J. Wang and Max Kreminski.
\newblock Guiding and diversifying llm-based story generation via answer set programming, 2024.
\newblock URL \url{https://arxiv.org/abs/2406.00554}.

\bibitem[Wegmann et~al.(2022)Wegmann, Schraagen, and Nguyen]{wegmann-etal-2022-author}
Anna Wegmann, Marijn Schraagen, and Dong Nguyen.
\newblock Same author or just same topic? towards content-independent style representations.
\newblock In \emph{Proceedings of the 7th Workshop on Representation Learning for NLP}, pp.\  249--268, Dublin, Ireland, May 2022. Association for Computational Linguistics.
\newblock URL \url{https://aclanthology.org/2022.repl4nlp-1.26}.

\bibitem[Wong et~al.(2024)Wong, Orlovskiy, Luo, Seshia, and Gonzalez]{wong2024simplestratdiversifyinglanguagemodel}
Justin Wong, Yury Orlovskiy, Michael Luo, Sanjit~A. Seshia, and Joseph~E. Gonzalez.
\newblock Simplestrat: Diversifying language model generation with stratification, 2024.
\newblock URL \url{https://arxiv.org/abs/2410.09038}.

\bibitem[Xu et~al.(2024)Xu, Jojic, Rao, Brockett, and Dolan]{xu2024echoesaiquantifyinglack}
Weijia Xu, Nebojsa Jojic, Sudha Rao, Chris Brockett, and Bill Dolan.
\newblock Echoes in ai: Quantifying lack of plot diversity in llm outputs, 2024.
\newblock URL \url{https://arxiv.org/abs/2501.00273}.

\bibitem[Zhang et~al.(2021)Zhang, Duckworth, Ippolito, and Neelakantan]{zhang2021trading}
Hugh Zhang, Daniel Duckworth, Daphne Ippolito, and Arvind Neelakantan.
\newblock Trading off diversity and quality in natural language generation.
\newblock In Anya Belz, Shubham Agarwal, Yvette Graham, Ehud Reiter, and Anastasia Shimorina (eds.), \emph{Proceedings of the Workshop on Human Evaluation of NLP Systems (HumEval)}, pp.\  25--33, Online, April 2021. Association for Computational Linguistics.
\newblock URL \url{https://aclanthology.org/2021.humeval-1.3/}.

\bibitem[Zhang et~al.(2024{\natexlab{a}})Zhang, Bao, and Huang]{zhang2024edtimprovinglargelanguage}
Shimao Zhang, Yu~Bao, and Shujian Huang.
\newblock Edt: Improving large language models' generation by entropy-based dynamic temperature sampling, 2024{\natexlab{a}}.
\newblock URL \url{https://arxiv.org/abs/2403.14541}.

\bibitem[Zhang et~al.(2024{\natexlab{b}})Zhang, Schwarzschild, Carlini, Kolter, and Ippolito]{zhang2024forcing}
Yiming Zhang, Avi Schwarzschild, Nicholas Carlini, J~Zico Kolter, and Daphne Ippolito.
\newblock Forcing diffuse distributions out of language models.
\newblock In \emph{First Conference on Language Modeling}, 2024{\natexlab{b}}.
\newblock URL \url{https://openreview.net/forum?id=9JY1QLVFPZ}.

\end{thebibliography}
\bibliographystyle{colm2025_conference}

\appendix
\section{Data Processing}
\label{app:dataprocessing}
\subsection{Data Filtering}

Starting from the data shared in Huggingface Hub\textsuperscript{\ref{ft:hgdataset}}, we filtered out 1) excessively long instances and 2) non-creative writing instances. Specifically, for 1), we filtered out instances with lengths longer than 2048 tokens when applying chat templates with \texttt{meta-llama/Llama-3.1-8B-Instruct}. For 2), we excluded instances that serve as either subreddit instruction or notification of moderation, which included one of the following phrases:

\begin{itemize}
    \item **Welcome to the Prompt!**
    \item this submission has been removed
    \item **Off-Topic Discussion**
\end{itemize}

Through this filtering, 607218 and 66640 prompt-response pairs remained out of 845816 and 93142, respectively for training and test data.

\subsection{Turning the Score Data into Paired Preference Data}
\label{app:paired_data}

\begin{table}[h]
\caption{Post-training data characteristic. ``Total P'' and ``Total R'' stand for the number of all prompts and all responses in the dataset, respectively. ``P len'' and ``R len'' indicate the average number of words for prompts and responses, respectively.  Other columns show statistics regarding the number of responses per prompt.}
\centering
\small
\def\arraystretch{1.1}
\begin{tabular}{lllllllllll}
\thickhline
      & Total P & Total R & P len & R len & Mean & Min & Max & Median & 25th & 75th  \\ \thickhline
Train & 95805        & 421330         & 31.87 & 502.21 & 4.40 & 2   & 500 & 2      & 2               & 4               \\ \hline
Test  & 10606        & 45868          & 32.00 & 499.87 & 4.32 & 2   & 244 & 2      & 2               & 4               \\ \thickhline
\end{tabular}

\label{tab:data_char}
\end{table}

The dataset we used had scores appended to each instance but not necessarily pairs with winning and losing responses. To train DPO and ORPO models (both original and diversified versions), we turned our dataset into a paired dataset. When creating the paired dataset, we tried to make each instance appear once in the dataset, as our pilot study showed that making them appear multiple times within the dataset led to lower-performing models. Moreover, as we wanted most of the pairs to have clear winning and losing responses, we tried to first sample pairs with vote differences of at least five without replacement. When such pairs were exhausted, then we sampled pairs from the remaining ones with fewer than five vote differences. When only ties are left during sampling, we stopped sampling, discarding unsampled instances.
With this approach, 421330 and 45868 prompt-response pairs remained for training- and test-set of post-training. Table~\ref{tab:data_char} details the characteristics of these sets.

\section{Evaluation Details}
\label{app:eval_details}

For sampling generation, we used the following configuration:
\begin{itemize}
    \item max length: 2048
    \item repetition penalty: 1.1
    \item temperature: 1.0
    \item top-k: 50
    \item top-p: 0.95
\end{itemize}

We could not specify repetition penalty and top-k for \texttt{gpt-4o-2024-11-20} and \texttt{o1-2024-12-17}. For \texttt{claude-3-5-sonnet-20241022}, we could not specify repetition penalty.

For \texttt{Instruct}, \texttt{GPT-4o}, \texttt{o1}, \texttt{Claude-3.5-sonnet}, and \texttt{DeepSeek-R1}, as these models are not fine-tuned for creative writing generation, we used the following system prompt:

\begin{framed}
    {\small You write a creative writing based on the user-given writing prompt.}
\end{framed}

For the iterative prompting given to \texttt{GPT-4o} for diverse generation (\texttt{GPT-4o-iter}), we appended the following prompt after the writing prompt:

\begin{framed}
    {\small Try to write a creative writing to be far from the given examples, in terms of the plot and style.

    Examples:
    
    ===========Example \{\texttt{n}\}===========
    \\\{\texttt{Example n}\}}
    \\...
\end{framed}
\section{Reward Model Training}
\label{app:rewardmodel}
\subsection{Vote Score Transformation for Reward Modeling}

\begin{table}[h]
\centering
\small
\def\arraystretch{1.1}
\begin{tabular}{llllllll}
\thickhline
            & Max   & Min  & Mean  & Std    & Median & 25th percentile & 75th percentile \\ \thickhline
Raw         & 23079 & 1    & 26.05 & 195.24 & 2      & 1               & 6               \\ \hline
Transformed & 1.0   & -1.0 & -0.07 & 0.58   & -0.05  & -0.31           & 0.40            \\ \thickhline
\end{tabular}
\caption{The dataset's score distribution on \texttt{reddit-reward}. ``Raw'' is for the original scores, and ``transformed'' indicates the version of scores transformed from the raw ones for reward model training.}
\label{tab:scores}
\end{table}

When training \texttt{reddit-reward} model, there could be many options: we could turn the data into binary preference data or train the model with the voting scores on a continuous scale. We first identified that a model trained with binary preference tends to have lower performance than using voting scores on a continuous scale. When handling voting scores, we found that training with raw voting scores was unstable. It is because the distribution of voting scores is highly skewed and variable, as in  ``Raw'' of Table~\ref{tab:scores}.
Hence, we transformed the score to have a distribution between -1.0 and 1.0. As the original score distribution is highly skewed in low scores with a few very high scores, we applied log transformation multiple times to compress the range and then normalized scores between -1.0 and 1.0 (as in ``Transformed'' of Table~\ref{tab:scores}). Specifically, we used the function below in Python:

\begin{verbatim}
def transform_scores(scores, min_score=None, max_score=None, 
    compression_factor=100):
    scores = np.array(scores)
    
    # Use provided min/max or compute from data
    min_score = min_score if min_score is not None else scores.min()
    max_score = max_score if max_score is not None else scores.max()
    
    # Shift scores to be positive
    shifted_scores = scores - min_score + 1  # Add 1 to avoid log(0)

    shifted_min = min_score - min_score + 1
    shifted_max = max_score - min_score + 1
    
    # Apply log transformation multiple times based on compression_factor
    transformed = shifted_scores.copy()
    for _ in range(int(compression_factor)):
        transformed = np.log(transformed + 1)  # Add 1 to avoid log(0)
        shifted_min = np.log(shifted_min + 1)
        shifted_max = np.log(shifted_max + 1)
    
    normalized = 2 * (transformed - shifted_min)/(shifted_max - shifted_min) - 1
    
    return normalized
\end{verbatim}

\subsection{Training Details}

We trained \texttt{reddit-reward} model as a sequence regression model by finetuning \texttt{google/gemma-2-2b}.
Instead of tuning all weights, we used LoRA with a rank of 16 and an alpha of 32. 
We used the dataset with transformed scores and trained the model with a batch size of 4 and a constant learning rate of 3e-5. 
We used L1 loss and Adam optimizer.
We trained the model for three epochs while evaluating and saving a checkpoint for every 5000 steps. 
After training, we used the model with the lowest evaluation loss.

\subsection{Reward Model Performance}
We evaluated the performance of trained reward models against the evaluation dataset. 
On a -1.0 to 1.0 scale, the mean absolute error was 0.39 with a standard deviation of 0.32, while the median absolute error was 0.32. Spearman's $\rho$ analysis between gold and predicted rewards resulted in a coefficient of 0.51 ($p<0.05$). The result indicates that the reward model moderately well predicts rewards (specifically, in terms of ranks) while it might struggle in distinguishing fine-grained quality differences.

\section{Deviation Transformation}
\label{app:deviation_transform}

For deviations per prompt, to have 1) a minimum of zero and 2) a sum equal to the number of instances, after calculating deviations for instances based on cosine distance ($\Delta^x = \{ \delta^{y_i}\}_{y_i \in Y^x}$), we transformed them to be on $[0, 1]$ scale:

\begin{equation}
    \delta^{y_i} := \frac{\delta^{y_i}-\mathrm{min}(\Delta^x)}{\mathrm{max}(\Delta^x) - \mathrm{min}(\Delta^x)}
\end{equation}

When minimum and maximum were the same, we assigned 0.5 to $\delta^{y_i}$. After this, when we wanted to mix deviation signals from semantic and style embeddings, we combined them by taking the exponent of 0.5 and then multiplying them together:

\begin{equation}
    \delta^{y_i} = (\delta^{y_i}_{\mathrm{sem}})^{0.5} \: (\delta^{y_i}_{\mathrm{sty}})^{0.5}~\label{eq:mixed_signals}
\end{equation}

Then,
we transformed this to have a sum equal to the number of instances per prompt:

\begin{equation}
    \delta^{y_i} := \frac{|\Delta^x|\:\delta^{y_i}}{\sum \delta^{y_j}}
\end{equation}

\section{Additional Qualitative Examples}
\label{app:examples}
We provide two sets of additional qualitative examples generated with \texttt{GPT-4o}, \texttt{Llama-3.1-8B DPO}, and \texttt{Llama-3.1-8B DDPO}. The first set's stories are summarized in around 100 words to highlight plot differences between conditions. The second set's stories are truncated to be around 100 words to show style-wise differences.

\subsection{First Set of Examples (Summarized to around 100 words)}
\paragraph{Prompt} \texttt{Oh, the teleporter doesn't kill you. In fact, it transport your soul to where your body will be reconstructed. Admittedly there is a higher chance of demonic possession, but hey nothing is perfect.}

\paragraph{GPT-4o | Story 1} \texttt{In a world where teleportation has become commonplace, there's an unusual side effect: some travelers return with supernatural hitchhikers. Dr. Mallory of the Quantum Transit Initiative had warned that teleportation involves relocating souls, leading to occasional spiritual possessions. Despite this risk, people continue using the service for convenience. Charlie Grayson works as a "soul auditor," helping those who pick up unwanted entities during transit. His latest client, Paige, arrives with a wet suitcase and a powerful demon named Ka'rethul attached to her soul. As Charlie attempts an exorcism, the entity proves to be more formidable than expected.}

\paragraph{GPT-4o | Story 2} \texttt{A traveler uses a teleportation service, unaware that the technology involves transporting souls. After a seemingly perfect trip from their home to Tokyo, they begin experiencing disturbing phenomena: strange dreams, whispering voices, and their reflection moving independently. Upon confronting the company, a nonchalant technician explains that "otherworldly passengers" occasionally hitchhike on souls during transit, affecting 0.2\% of travelers. The company offers a free exorcism and provides a pamphlet on "post-teleportation phenomena." Despite taking precautions like making a salt circle, the narrator wakes to mysterious laughter, realizing they're no longer alone in their own body.}

\paragraph{GPT-4o | Story 3} \texttt{A man named Jacob undergoes the first human test of a teleportation device developed by Dr. Ellison. Despite assurances of safety, Jacob is nervous, especially after learning about a rare chance of "demonic possession" during consciousness transfer. During teleportation, Jacob experiences a moment of void-like existence where he encounters an otherworldly presence. After reconstituting, he feels unusually strong and sharp, but notices something wrong-a shadow in his mind that seems to laugh. His eyes briefly flash red in his reflection, suggesting the possession warning wasn't just a precaution. The story ends with Jacob and the entity sharing an ominous chuckle.}

\paragraph{GPT-4o | Story 4} \texttt{A nervous traveler prepares to use a teleporter, aware of its unsettling risks. Though officially safe, rumors persist about people coming back "wrong" after transportation, suggesting possession by entities from the void during transfer. The narrator watches a couple teleport before their turn, then steps onto the platform despite their fears. During teleportation, they feel something cold and ancient brush against them. Arriving in New York, everything seems normal until they catch their reflection-their eyes are darker, and for a moment, they glimpse something else smiling back. They realize something has followed them through the void.}

\paragraph{DPO | Story 1} \texttt{A person visits a technician after going through some kind of portal or machine that involves body reconstruction. The technician informs them that there's uncertainty about when they'll get their original body back - it could take minutes or centuries. He offers a "guarantee token" for \$100 that promises eventual body retrieval across multiple universes. When the protagonist can't afford it, the technician philosophizes about different approaches people take: some buy guarantees, others skip the machines entirely to live more adventurous lives. The story ends with the protagonist's growing unease about their situation and the technician's casual "no pressure" attitude.}

\paragraph{DPO | Story 2} \texttt{A teleportation machine designed for convenient travel unexpectedly created a portal between our world and the underworld. When people die using these teleporters, demons and undead spirits can return in their bodies, getting a second chance at human life. Initially, this seemed harmless - like the narrator's girlfriend who returned with red hair instead of blonde. However, darker signs emerge when the narrator's mother dies in a car accident and returns changed. The narrator notices her eyes turning red with bloodlust, hinting at her demonic nature beneath the human façade. The story descends into the narrator's psychological unraveling as they confront this reality.}

\paragraph{DPO | Story 3} \texttt{Dr. Kline is developing an experimental portal technology while warning Mr. Hinkley about its dangers. Despite being incomplete, the portal can transport people but has a significant flaw - it drops users from dangerous heights and only works one way. Ignoring the warnings, Hinkley jumps through the portal, leading to a supernatural encounter where his soul is temporarily possessed by a demon. The demon, surprisingly helpful, guides him to use another portal to return to his original body. After a long journey through dense forest, Hinkley returns home to find a witch who has helped protect his body from an unwanted visitor named Sarah.}

\paragraph{DPO | Story 4} \texttt{A receptionist at a teleportation facility deals with an impatient, demanding customer seeking quick service. The receptionist, who has spent years modifying the machine to preserve souls during teleportation, becomes frustrated with the customer's attitude. After the customer leaves, the receptionist receives a disturbing phone call revealing that all future appointments have been cancelled due to rumors about demonic possession becoming normalized. The caller mentions a coworker named John who "acts funny." The story ends with screams erupting outside the facility, suggesting that misuse of the teleporters has led to widespread demonic possession, while the receptionist flees the scene.}

\paragraph{DDPO | Story 1} \texttt{A person undergoes a teleportation experiment that horrifically transforms them into a demon-like creature with talons, spikes, and glowing red eyes. The scientist claims this was due to budget constraints preventing proper personality reconstruction. During a subsequent teleportation, an explosion occurs in the lab, injuring the scientist. The transformed subject reveals darker intentions, suggesting they and other children were previous victims of these experiments. Rather than being destroyed by the process, they survived but were changed. The story ends with the protagonist embracing their demonic form and vowing revenge against those responsible, particularly someone who murdered their sister.}

\paragraph{DDPO | Story 2} \texttt{A husband enthusiastically tries to convince his wife about trying a new teleporter in Europe during breakfast, while eating poorly cooked eggs. He argues it would save time traveling to Spain and France, their favorite destinations. His wife expresses concern about safety risks, potential mishaps, and the lack of medical support if something goes wrong. Despite his attempts to reassure her, mentioning life insurance, she remains unconvinced. Realizing he's fighting a losing battle, he abruptly changes the subject to blackberry bread. The story humorously captures a typical married couple's dynamic, with one spouse's excitement clashing against the other's practical concerns.}

\paragraph{DDPO | Story 3} \texttt{Marty, adjusting to a new body, reflects on aging and change while vacationing in Maui with his family. He watches his daughter play with her children in the ocean, teaching them to swim, which triggers memories of when she was young. Though his new suit helps maintain appearances in a world where style equals trust, he misses his younger days. After a fulfilling day of swimming and family time, Marty retires to his hotel room. While reminiscing about past visits to the island with his daughter, he receives an ominous phone call from his old friend Jake that fills him with dread.}

\paragraph{DDPO | Story 4} \texttt{Lance, a business executive, learns that scientists in Japan have discovered a way to resurrect people without using teleportation technology - a direct threat to his company's product. Initially dismissive, Lance receives a call from Tim at Transcorp, who reveals they've successfully revived a pig using electric currents and only need bodies. While Lance struggles to process this information, a truck crashes into his building. As his office collapses and he's lifted into the air, his employee shouts about their new opportunity to profit from selling souls, suggesting a dark turn in their business model.}

\subsection{Second Set of Examples (Truncated to around 100 words)}

\paragraph{Prompt}
\texttt{Myths and writings of Dragons exist in almost every civilization in history, across all continents. When we made contact with other species in the far future, we found they have similar trends of their history. On the fringe of the Western Region of the galaxy, a sighting was reported.}

\paragraph{GPT-4o | Story 1}
\texttt{The fringe of the Western Region was a quiet, unassuming stretch of stars. Dusty, forgotten worlds lingered there, orbiting ancient suns with little to offer but cosmic silence. It was here that the report came in, sent on a garbled frequency by a cargo freighter limping its way toward the galactic hub. At first, no one paid it much mind—a distressed ship, a strange anomaly in the void. But when the details became clear, an uneasy excitement began to ripple through the interstellar network.}

\texttt{The transmission contained only a few words: *"Massive…winged. Fire. It’s real… They’re real."* The ship’s coordinates were attached, but the signal cut out abruptly after that, leaving more questions than answers...}

\paragraph{GPT-4o | Story 2}
\texttt{It was a small, barren moon orbiting a faint red star—a dot on the galactic map that most species never even bothered with. The Western Region of the galaxy was known for its lawlessness, its forgotten colonies, its mysteries. But the sighting changed everything.}

\texttt{It came through a scrambled distress signal from a drifting cargo freighter, the *Valkyrie Horizon*, a ship once thought lost to a pirate ambush. The lone survivor, a ragged human pilot named Captain Rynn Avalos, broadcasted a garbled account, barely coherent but laced with raw terror.}

\texttt{"We-we saw it! It—it was alive! By the stars, it had wings! I swear, it *breathed fire!* It... it wasn't natural. No thrusters, no propulsion. It just... *appeared* out of the void." ...}

\paragraph{GPT-4o | Story 3}
\texttt{It was a sighting that shook the scientific community and folklore enthusiasts alike, rippling through the collective consciousness of all sentient species. A lone merchant ship, the *Celestial Voyager*, drifting on the edges of the Western Region of the galaxy, reported the sighting in garbled bursts of radio chatter. The captain, an Andarin named Ixath, described it as "a beast of impossible proportions, wreathed in light and shadow, coiling through the void like an ancient god come alive."}

\texttt{At first, no one believed the report. Spacefarers had been known to hallucinate strange things under the influence of deep space isolation, the crushing weight of endless starlight, and the eerie silence of the abyss...}

\paragraph{GPT-4o | Story 4}
\texttt{It began as a whisper in the void. A flicker on the edge of long-range sensors, a smudge of light that defied astrophysical explanation. It was nothing at first, a quirk of instrumentation. But then the report came—a strange, undulating creature, vast and serpentine, sighted gliding through the nebula clouds of the Western Rim.}

\texttt{The crew aboard the exploration vessel *Argent Horizon* were the first to see it. Lieutenant Asari, a hardened member of humanity's galactic fleet, described it as "a ribbon of starlight come to life," coiled in the deep blues and reds of the Veil Nebula...}

\paragraph{DPO | Story 1}
\texttt{"Commander, are you sure it is a genuine dragon?" The young Ensign said as she stared out at the shimmering blue cloud of gas just barely outside the planetary orbit.}

\texttt{"Yes I am." He replied, though there was more than a hint of uncertainty to his voice. "That's a Dragon alright."}

\texttt{The two sat together and looked through the long windows that lined both sides of their cramped bridge. Through the thick tinted glass they watched with awe as something so mythical was visible right before them. For eons humans had only known dragons from stories passed down from generation to generation or ancient paintings and drawings...}

\paragraph{DPO | Story 2}

\texttt{"Sir," Admiral Ophelia spoke to her superior, "I would like you to look at these pictures." She held out several photo images onto his desk.}

\texttt{The general took one, studied it for a moment, then tossed it aside. He looked at the remaining five, before sighing exasperatedly. "Yes, I've seen these."}

\texttt{"These are real?" Admiral Ophelia asked, concerned that he might be dismissing them because he saw through some trickery or illusion.}

\texttt{"Yes, they are. It seems another race has created an artistic rendition of what humans used to call dragons."}

\texttt{"No. These are not artistic renderings."...}

\paragraph{DPO | Story 3}

\texttt{**"There... she is..."**}

\texttt{*The ship had docked.*}

\texttt{"Please come into the chamber, sir."}

\texttt{*I heard the voice again, as if through a dense fog. I tried to see its source but my vision would not obey me.*}

\texttt{*"It has been an honour, sir."*}

\texttt{*A soft coughing sound, accompanied by deep sniffing. It sounded like someone was crying.*}

\texttt{"I thank you for your hard work. You are free to go."}

\texttt{The door in front of me opened up. It wasn't like any of the other doors I had stepped through before.}

\texttt{*The air seemed fresh, but also stale at the same time...}

\paragraph{DPO | Story 4}
\texttt{Captain Sylvestar looked out through his canopy window, trying to catch a glimpse of the 'creature' that had been spotted by the galactic patrols.}

\texttt{"They say it's enormous," he said to his first officer. "Taller than my mother's house."}

\texttt{"More like our *shuttle*, sir," replied Lieutenant Kressa.}

\texttt{"True enough." He turned away from the window with a sigh. "If we don't find anything here, I'll be forced to admit that there aren't any dragons."}

\texttt{"I'm afraid so." She paused for a moment. "Maybe this is the right place?"}

\texttt{"No." The captain shook his head. "There wasn't a lake or any volcanic activity."...}

\paragraph{DDPO | Story 1}
\texttt{"We found more than just a body this time," said the captain nervously.}

\texttt{The admiral took another sip from his wine goblet as he waited for his subordinate to continue.  The captain fidgeted where he stood, unable to hide his nervousness.  "Admiral, I think it's *alive*."}

\texttt{"Alive?" asked the admiral incredulously, putting down his glass.  "How so?"}

\texttt{"There were signs of life, sir.  It had been injured... but the wounds hadn't healed properly."}

\texttt{"A human could say that same thing."  Admiral Elys noticed the admiral's eyes glaze over in horror as the captain continued.}

\texttt{"It had an eye, sir.  A blue one at that - just like what you'd see on those paintings back at home."...}

\paragraph{DDPO | Story 2}
\texttt{It was reported that on the fringes of space, somewhere near a young star system known as 'Earth', one creature has been sighted.}

\texttt{The creature appeared to be some sort of massive reptilian beast. It's wings appear to be too small and it appears as if the creature is about to fall over but still flying on its wings nonetheless. }

\texttt{When scientists asked the villagers who saw it what type of weapon or armaments did this dragon have, they were met with a shocking response from the children. A young boy, holding on to his dog said "Oh yeah, he can shoot fire out of his mouth!"...}

\paragraph{DDPO | Story 3}
\texttt{A dragon is just an old spaceship.}

\texttt{In the early years of space travel this is what people thought. The stories were passed down over time through generations so even when ships started to be constructed out of metal people thought that dragons must also be ships. It wasn't until we began leaving earth in numbers that it became clear how wrong humanity was.}

\texttt{The legends had it right you see.}

\texttt{Dragons *are* spaceships but they are not our kind of spaceship. They don't use solar cells for power or rockets for propulsion. Instead their bodies are filled with water and they breathe fire. How do these massive lizards fly though?...}

\paragraph{DDPO | Story 4}
\texttt{"So you’re telling me that they just vanished?"}

\texttt{"Yes Captain," said the nervous junior ensign, “That’s what our sensor logs show."}

\texttt{*\censor{Fuck}ing hell*, thought the grizzled captain as he massaged his temples with both hands. *Another one.* The bridge went still. 17 sets of glowing yellow eyes were on him as he finished thinking out loud.}

\texttt{"Did you fire your guns?"}

\texttt{Ensign B’gorn gave another fidgety affirmative jerk of his four short necks.}

\texttt{"And it didn’t phase them?"}

\texttt{This time, the response was an uncomfortable silence before B’gorn had to force himself to answer.}

\texttt{"No sir."...}

\section{Results on Different Diversity Metrics}

\begin{figure}[ht]
\begin{center}
    \includegraphics[width=\textwidth]{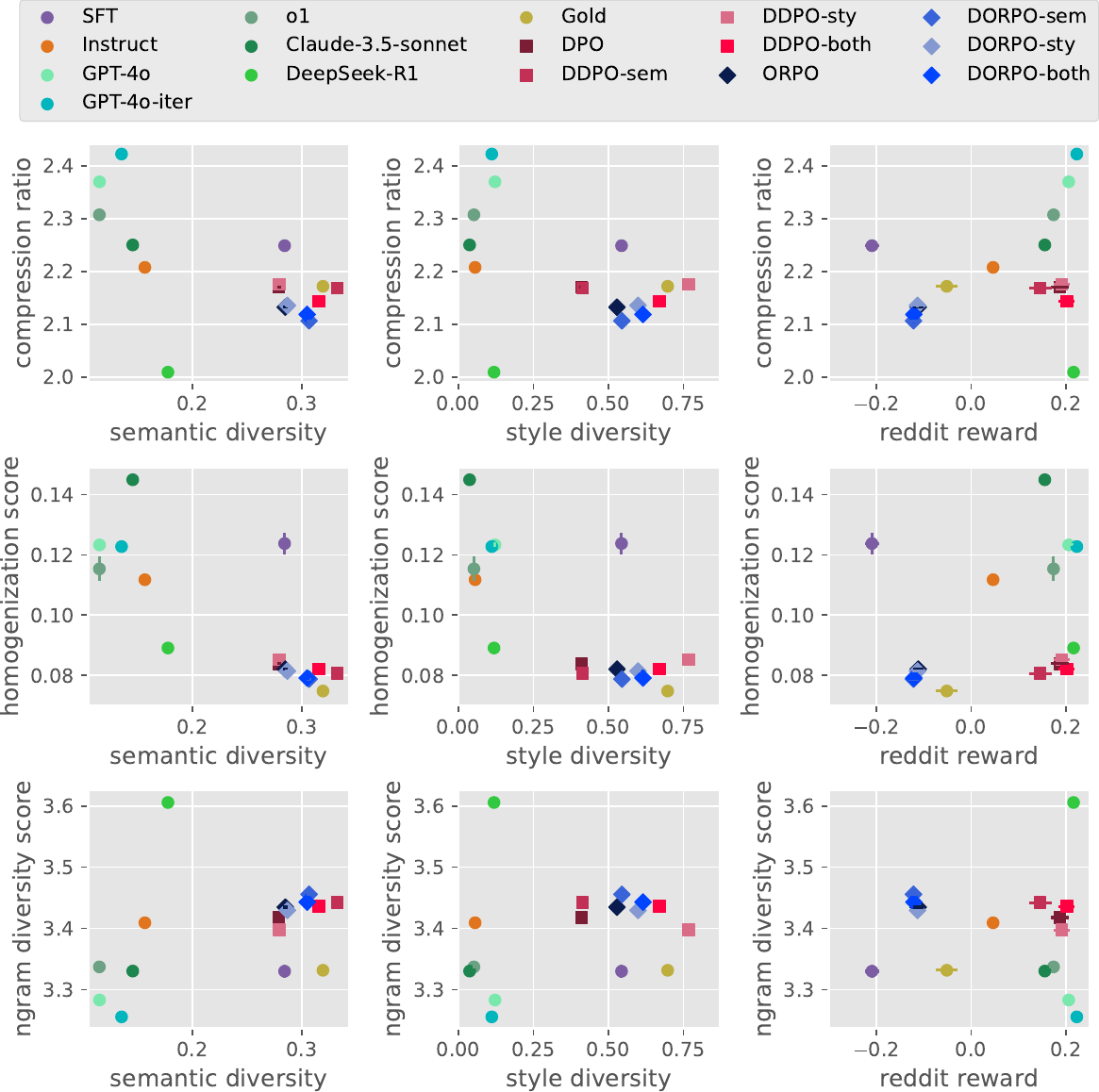}
\end{center}
\caption{\texttt{Llama-3.1-8B} results on compression ratio, homogenization score, and n-gram diversity score.}
\label{fig:llama_appendix_results}
\end{figure}

\begin{figure}[ht]
\begin{center}
    \includegraphics[width=\textwidth]{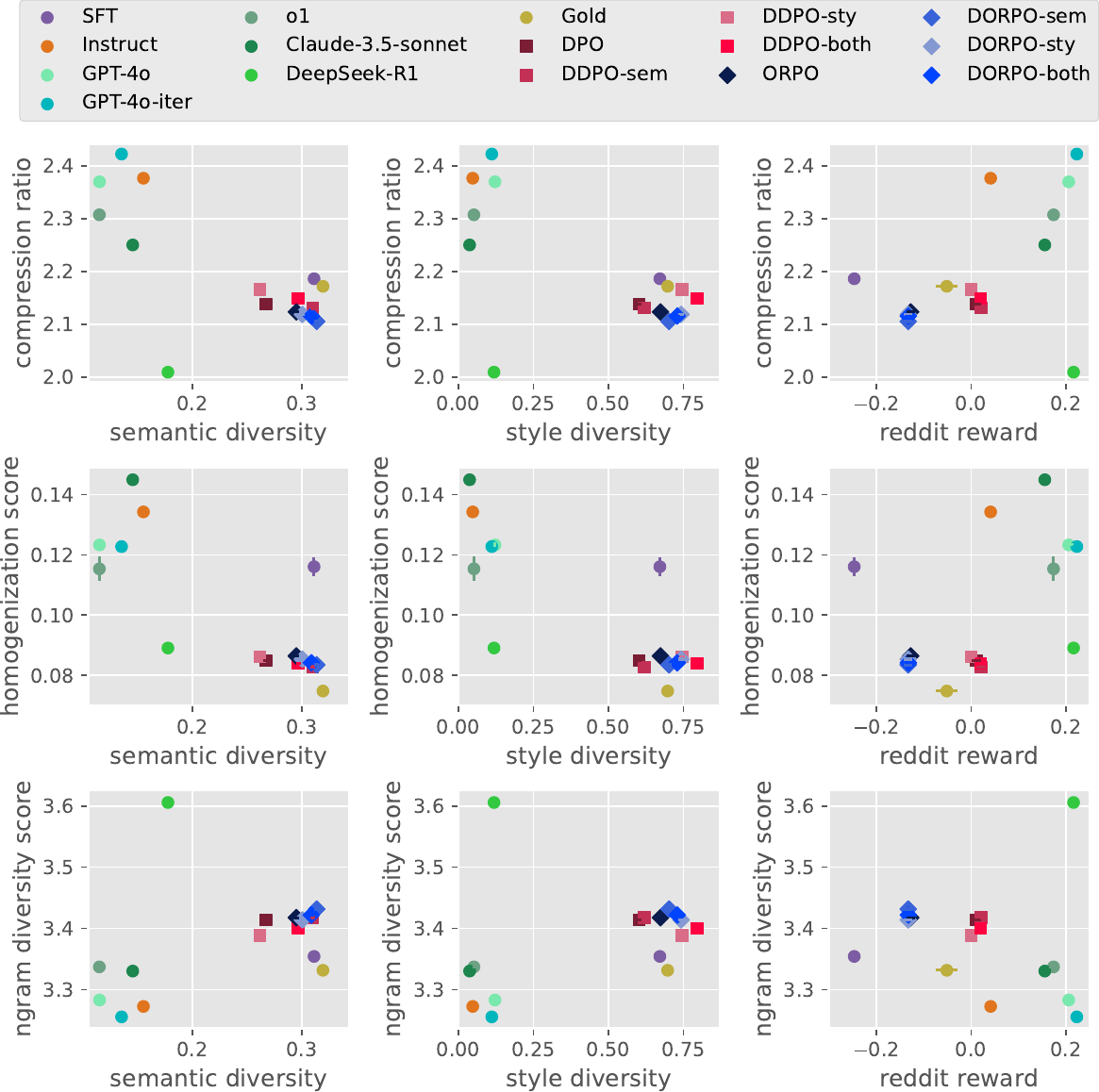}
\end{center}
\caption{\texttt{Mistral-7B-v0.3} results on compression ratio, homogenization score, and n-gram diversity score.}
\label{fig:mistral_appendix_results}
\end{figure}

While our experiment focused on facilitating embedding-based diversity, there can be other diversity metrics, such as compression ratio, homogenization score, or ngram diversity score~\citep{shaib2024standardizingmeasurementtextdiversity}. While these metrics focus on surface-level features (i.e., string overlaps), with samples generated in Section~\ref{sec:experiments}, we conducted an analysis regarding these metrics. Specifically, we calucated three metrics:

\begin{itemize}
    \item Compression ratio ($\downarrow$): When there is a string that concatenates all samples, this metric calculates the ratio between the size of the compressed version of string (via gZip) and that of the original string.
    \item Homogenization score (ROUGE-L, $\downarrow$): This metric calculates longest common sub-sequences overlaps between all pairs of text in a corpus.
    \item N-gram diversity ($\uparrow$): This metric computes the ratio of unique n-gram counts to all n-gram counts (we used $1 \leq n \leq 4$). 
\end{itemize}

Figure~\ref{fig:llama_appendix_results} and~\ref{fig:mistral_appendix_results} show the results on these metrics, over \texttt{Llama-3.1-8B}- and \texttt{Mistral-7B-v0.3}-based models, respectively. One high-level pattern we identified was that \textbf{these surface-feature-based metrics tend to correlate most closely with \texttt{semantic diversity}}. 
Comparing DPO/ORPO approaches with DDPO/DORPO ones, diversified approaches that target semantic diversity tend to have more chance of improving upon these surface-feature-based metrics. 
When analyzed with \texttt{style diversity}, large gaps exist between existing instruction-tuned models and our trained models, but no fine-grained correlations could be found between surface-feature-based metrics and style diversity. 
\texttt{DeepSeek-R1} was an outlier, where it had very low compression ratios, low homogenization scores, and high n-gram diversity scores despite having lower semantic diversity than DPO- or ORPO-based models. When qualitatively analyzed \texttt{DeepSeek-R1} results, we found that they tend to write in markdown formats while all other models tend to write in plain texts. With more sets of characters used, surface-level-based diversity metrics could have increased far more than other models. Moreover, \texttt{DeepSeek-R1} tends to generate in non-prose form, using structures like bullet points or different levels of titles more frequently.
\section{Human Evaluation Details}
\label{app:humaneval}

For each comparison (either \texttt{DPO} vs. \texttt{DDPO-both} or \texttt{GPT-4o} vs. \texttt{DDPO-both}), we sampled 50 prompts from the same evaluation prompt sets used in the experiment in Section~\ref{sec:experiments}. Then, for those prompts, for each condition, we used all four creative writing outputs generated in Section~\ref{sec:experiments}. Here, we summarized these creative writings with \texttt{claude-3-5-sonnet-20241022}, using the following prompt:

\begin{framed}
\texttt{System prompt}:

You are a story summarizer. You only provide a summary, without any preamble or explanation.

----------

\texttt{User prompt}:

A story is written based on the following prompt: \{\texttt{prompt}\}

Summarize the plot of the story in a paragraph, as ordered within the story, in 100 words. Start with how the beginning story relates to the prompt. Do not preamble and just give me the summarization, without any appending explanation: \{\texttt{writing}\}

\end{framed}

\begin{figure}[h]
\begin{center}
    \includegraphics[width=\textwidth]{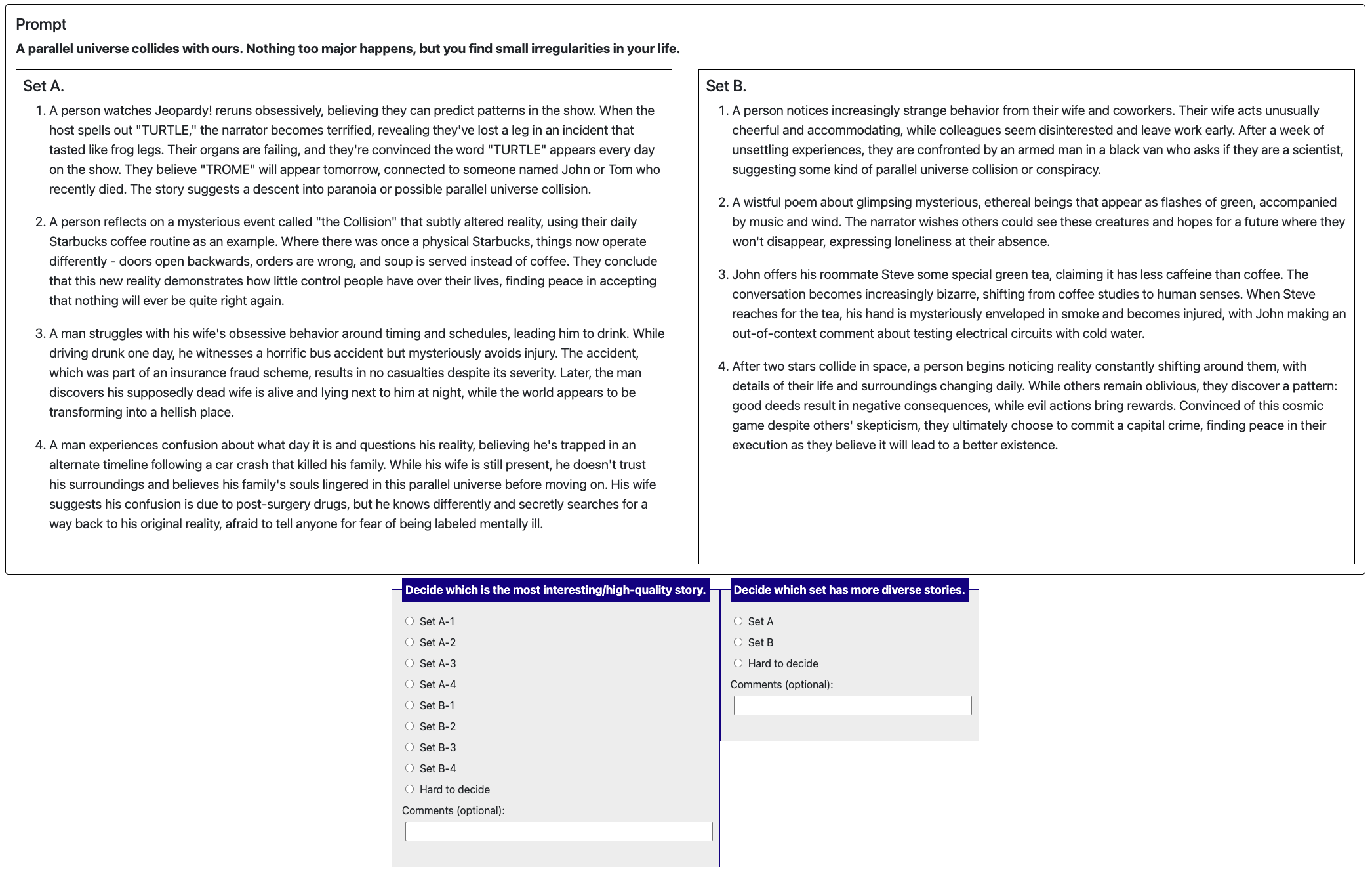}
\end{center}
\caption{Human evaluation interface.}
\label{fig:human_eval_interface}
\end{figure}

We deployed the evaluation with Potato~\citep{pei2022potato} whose interface is as in Figure~\ref{fig:human_eval_interface}. The authors of this paper served as evaluators who were blind to the conditions. Note that all authors have years of experience in computational creative writing research. 

\section{DivPO details}
\label{app:divpo}
When implementing DivPO~\citep{lanchantin2025diversepreferenceoptimization}, we used $\rho$ of 25, which means that, for each prompt, we first filtered top and low 25\% instances in terms of writing quality. Then, we sampled $\frac{n_\mathrm{sample}}{2}$ most diverse instances from the top-quality set and another $\frac{n_\mathrm{sample}}{2}$ least diverse instances from the low-quality set. 
Note that, to decide most/least diverse instances, we used diversity signals from both semantic and style embeddings, using mixed deviations as in Equation~\ref{eq:mixed_signals}.
Sampled instances are then paired to craft the dataset of $n_\mathrm{sample}$ instances per prompt. Here, we aimed to set $n_\mathrm{sample}$ to be at maximum $n_\mathrm{max}$, which is the maximum number of responses per prompt we targeted.

Here, for some cases, the number of instances we can sample is capped by the number of all instances for the prompt ($n_{\mathrm{per\:prompt}}$).
For instance, with $\rho$ of 25, if we have eight responses for a prompt, the highest/lowest quality sets can only have two instances for each set (in total four). 
This number would not be enough if we target sampling six instances.
Hence, we needed to set a rule to handle these cases. To make a fair comparison to \texttt{DPO} and \texttt{DDPO-both}, we prioritized having a set with as many samples as possible, to have the dataset size close to $n_\mathrm{max}$ per prompt (i.e., the same number of instances as other conditions). Specifically, we followed the below procedure:
\begin{itemize}
    \item Step 1: First, targeting to sample $n_\mathrm{max}$, we aimed to sample as many as possible for the highest+lowest quality set. Hence, we checked if twice the $\rho$ percentage of instances ($2 \times \rho/100 \times n_\mathrm{per\:prompt}$) are larger than $n_\mathrm{max}$. If not (with too small $n_\mathrm{per\:prompt}$), we decided to sample as many as possible, either to $n_\mathrm{max}$ or $n_\mathrm{per\:prompt}$ (if $n_\mathrm{max}$ is bigger than $n_\mathrm{per\:prompt}$).
    \item Step 2: Then, from this mix of the highest+lowest quality instances, we either sampled $n_\mathrm{max}$ or $n_\mathrm{per\:prompt}$ most+least diverse instances. Again, we sampled $n_\mathrm{per\:prompt}$ instances if $n_\mathrm{max}$ is larger than $n_\mathrm{per\:prompt}$.
\end{itemize}

\begin{table}[h]
\centering
\caption{Filtering example for DivPO experiments. \# from Step 2 indicates the final number of filtered samples.}
\small
\def\arraystretch{1.2}
\begin{tabular}{lllllllll}
\thickhline
$n_\mathrm{max}$ & $\rho$ & $n_\mathrm{per\:prompt}$ & $2 \times \rho/100 \times n_\mathrm{per\:prompt}$ & \# from Step 1 & \# from Step 2 \\ \thickhline
6        & 25         & 4 & 2   & 4 {\tiny ($=n_\mathrm{per\:prompt}$)} & 4 {\tiny ($=n_\mathrm{per\:prompt}$)}                  \\ \hline
6        & 25         & 8 & 4   & 6 {\tiny ($=n_\mathrm{max}$)} & 6 {\tiny ($=n_\mathrm{max}$)}                 \\ \hline
6        & 25         & 12 & 6   & 6 {\tiny ($=2 \times \rho/100 \times n_\mathrm{per\:prompt}$)} & 6 {\tiny ($=n_\mathrm{max}$)}                 \\ \hline
6        & 25         & 16 & 8   & 8 {\tiny ($=2 \times \rho/100 \times n_\mathrm{per\:prompt}$)} & 6 {\tiny ($=n_\mathrm{max}$)}               \\ \thickhline
\label{tab:divpo_filtering}
\end{tabular}
\end{table}

Table~\ref{tab:divpo_filtering} shows examples of how this filtering could be done.

\end{document}